\documentclass[11pt]{article}
\usepackage[margin=1in]{geometry}
\usepackage[T1]{fontenc}
\usepackage{lmodern}
\usepackage{graphicx}
\graphicspath{{./}}
\usepackage{float}
\usepackage{amsmath,amssymb}
\usepackage{array}
\usepackage{booktabs}
\usepackage{multirow}
\usepackage[numbers,sort&compress]{natbib}
\usepackage{xcolor}
\usepackage[colorlinks=true,allcolors=blue!55!black]{hyperref}
\usepackage{caption}
\usepackage{tikz}
\usetikzlibrary{arrows.meta,decorations.pathreplacing,positioning,calc}
\newcommand{\Mistral}{Mistral-7B}
\newcommand{\Llama}{Llama-3.1-8B}
\newcommand{\LlamaBig}{Llama-3.1-70B}
\newcommand{\W}{W}
\newcommand{\band}{S}

\newcommand{\rd}[2]{#1/#2}
\newcommand{\llamacpp}{\mbox{llama.cpp}}

\makeatletter
\newcommand{\secbarrier}{\par\vspace{0pt}%
  \ifx\@deferlist\@empty\else\clearpage\fi}
\makeatother
\title{Retrofitting Unbounded Context onto Stock Language Models}
\author{Mark Oskin\\
Professor\\
School of Computer Science and Engineering\\
University of Washington\\
\texttt{mhoskin@uw.edu}}
\date{November 2026}

\begin{document}
\maketitle

\begin{abstract}
A language model that never had to forget would keep every document it has read and every turn it has spoken, and
answer from any of it. This paper shows that two stock models can do that on unmodified weights by manipulating their
geometry. Rotary attention turns each pair of key dimensions, a \emph{plane}, at its own rate. The slow planes turn
once in thousands of tokens and carry the address a query matches when it reads a far key. The fast planes turn once
in a few tokens, and the order of the tokens inside a span is already written into the span's keys and values by the
time the span leaves the near window. At distance the position blurs, and what attention reads is a bag rather than
a sequence.

Each attention head learns a function. A ``copy head'' finds an earlier occurrence of the current token and copies
from it, weighting a longer aligned match more. A head turns itself off by attending to a ``sink'' token. A nearby
match is positional; a far match is position free. Under competition, a far span is read only if it is visible at
the query of the question's own handle word. A read through a description rather than a word fails at distance:
description-matched reads that hold at 95~percent at the front of the window fall to 0--16~percent at 32K and 128K in
every model measured, a 70B included.

These facts allow a trick. What sits in the far part of attention does not have to be the history as it was; it can
be installed ahead of the prediction. The design keeps a near window of 8K tokens at their true positions and
changes what lies beyond it with the prompt: every far key and value goes to the host exact, a word index finds the
units the request's rare words name, a far window of slots inside the trained range holds them while the request is encoded, and a rare word the model generates fetches its units mid-generation with one exact re-encode. The same
geometry compresses the store: keeping 28 of 64 key planes at four bits loses no read on Mistral-7B, although the engine keeps the store uncompressed. Implemented in \llamacpp{} on Mistral-7B and Llama-3.1-8B, the design
reads planted single needles at 99--100 of 100 and three-key discrimination at 88--100 of 100 from 32K to one million
tokens. On RULER it scores 70.8 against 69.9 for dense attention on Llama-3.1-8B at 128K (a near window alone scores 14.8) and
67.8 against 75.5 on Mistral-7B at 32K. At 128K it prefills at 1.85 times and decodes at 1.65--1.68 times stock
speed. Two-hop questions reach parity with dense attention on Llama-3.1-8B (92 against 92 at 128K) and stay 25 points
below it on Mistral-7B at 32K (66 against 91), bounded by the first hop, which the near window alone loses.
\end{abstract}

\section{Introduction}
\label{sec:intro}

A language model is trained to attend over a fixed number of tokens, its context window (e.g. 32,768 tokens for \Mistral{}, 131,072 for \Llama{}). Everything the model can use about a document
must fit in that window, and a long document, a code base or a long conversation often does not. Training a model to
a longer window costs more in proportion, and so does serving it, since the memory attention keeps grows with the
window. We ask whether a model that was never trained beyond its window can read a history of any length anyway,
with no retraining and no change to its weights. This paper shows that two stock models can, and that the reason
lies in how attention already treats the distant part of its window.  The idea present here is generally applicable
to a wide class of language transformers.

Attention works by matching. For every token the model has read it keeps a key and a value; to produce the next
token it compares a query against every key and reads the values of the keys that match best. Position enters
through rotary encoding~\citep{su2021roformer}: each key is turned by an angle proportional to its position, in many
independent planes that turn at different rates. The fast planes turn once every few tokens and tell neighbors
apart; the slow planes turn once in thousands of tokens and tell regions of the document apart. We measured two
stock models, \Mistral{}~\citep{jiang2023mistral} and \Llama{}~\citep{grattafiori2024llama3}, and find that beyond a
near window of a few thousand tokens only the slow planes still matter. A distant token is matched by the words it
carries and by its rough region rather than by its exact position, so its key can be presented at any position
inside the trained window and still be found, and the order of the tokens inside a distant span was already written
into the span's keys and values while the span was near. The far part of the window is therefore a bag of spans
rather than a sequence, and a bag can be filled with whatever the model needs rather than with whatever came last.

What a far read needs is specific. It is lexical: a distant passage is found by a word it shares with the question,
and a passage that is only described, with no word in common with the question, is not found at distance by any
model we measured, a 70B model included, although descriptions do work near the question. It is bound at one moment:
when several passages of the same form compete, the one the model reads is the one present when the key word of the
question is encoded, so a passage fetched later is not read until the question is encoded again. And the values are
the payload: a wrong value is worse than no value at all. This behavior means we need not understand the concepts
behind a token to fill the far window with useful history; we need only match on the tokens themselves.

\begin{figure}[htbp]
\centering
\resizebox{\textwidth}{!}{%
\begin{tikzpicture}[
  x=1cm, y=1cm, font=\footnotesize,
  cell/.style={draw, rounded corners=1.5pt, align=center, inner sep=2.5pt, font=\scriptsize},
  lab/.style={font=\scriptsize, align=center, inner sep=1pt},
  row/.style={font=\footnotesize\bfseries, anchor=west, inner sep=1pt},
  arr/.style={-{Stealth[length=2.2mm]}, thick},
  num/.style={circle, draw, fill=white, inner sep=1.1pt, font=\scriptsize\bfseries}
]
\node[row] at (0,12.05) {(a) store};
\fill[gray!8] (0,10.55) rectangle (11.0,11.7);
\draw (11.0,10.55) -- (0,10.55);
\draw (11.0,11.7) -- (0,11.7);
\draw[dashed] (0,10.55) -- (0,11.7);
\node[cell, fill=blue!22, minimum width=2.1cm, minimum height=0.9cm, anchor=west] (unitA) at (0.8,11.125)
  {a sentence unit\\with a rare word $\star$};
\node[lab, anchor=north] at (1.85,10.47) {true position $p$};
\node[lab] at (6.95,11.125) {far history, at true positions:\\every token here has left the near window once};
\node[cell, fill=green!14, minimum width=4.0cm, minimum height=1.15cm, anchor=west] (winA) at (11.0,11.125)
  {near window\\the last 8,192 tokens\\at true positions};
\draw[arr] (15.05,11.125) -- (15.9,11.125);
\node[lab, anchor=south] at (15.5,11.2) {prefill};
\draw[thick] (11.0,10.55) -- (11.0,9.6);
\node[num] at (11.0,10.25) {1};
\draw[arr] (11.0,9.6) -- (5.0,9.6) -- (5.0,9.1);
\draw[arr] (11.0,9.6) -- (11.0,9.1);
\node[lab, anchor=south] at (6.9,9.68) {capture: keys and values of every layer, exact, to the store};
\node[lab, anchor=south west] at (11.3,9.68) {a whole word above 15 nats of surprisal\\enters the index with its sentence unit};
\node[row] at (0,9.35) {(b) retrieve};
\node[cell, fill=yellow!14, minimum width=7.0cm, minimum height=1.6cm, anchor=west] (store) at (0,8.3) {};
\node[cell, fill=blue!22, minimum width=2.0cm, minimum height=0.8cm, anchor=west] (unitS) at (0.25,8.3)
  {unit $\star$\\as written at $p$};
\node[lab, text width=4.5cm] at (4.7,8.3) {host store, unbounded: every far token's keys and values, all 32 layers, exact, as written; 131 KB per token};
\node[cell, fill=yellow!14, minimum width=4.1cm, minimum height=1.6cm, anchor=west, text width=3.8cm] (index) at (8.3,8.3)
  {word index: $\star \mapsto$ its last 16 occurrences, each a sentence unit (a word enters above 15 nats of surprisal)};
\draw[arr] (index.west) -- (store.east) node[midway, above=1pt, lab] {postings};
\node[lab, text width=3.6cm] at (14.55,8.3) {the host grows with the document; the GPU cache holds about 25K cells per layer at any length};
\draw[arr] (16.2,5.3) -- (16.2,6.95) -- (10.4,6.95) -- (10.4,7.5);
\node[num] at (11.6,6.95) {2};
\node[num] at (12.6,6.95) {4};
\node[lab, anchor=north] at (12.9,6.72) {2: the request's rare words and the seed;\\4: a rare word the model generates, then the\\request and tail are encoded again over it};
\draw[arr] (3.15,7.5) -- (3.15,5.3);
\node[num] at (3.15,6.5) {3};
\node[lab, anchor=north west] at (3.45,7.15) {placed at a slot $s$ just beyond\\the near window; keys re-rotated\\$p \to s$, plane by plane};
\node[row] at (0,5.6) {(c) insert};
\node[cell, fill=orange!18, minimum width=1.9cm, minimum height=1.3cm, anchor=west] (sink) at (0,4.65)
  {sink and\\pinned prefix\\129 slots};
\node[cell, fill=blue!6, minimum width=9.1cm, minimum height=1.3cm, anchor=west] (far) at (1.9,4.65) {};
\node[cell, fill=blue!22, minimum width=2.0cm, minimum height=0.85cm, anchor=west] (unitC) at (2.15,4.65)
  {unit $\star$\\at slot $s$};
\node[lab, text width=6.5cm] at (7.6,4.65) {far window (the slot band): 16,384 slots, always full with the fetched units, units kept from the last gather, and the most recent far tokens};
\node[cell, fill=green!14, minimum width=3.5cm, minimum height=1.3cm, anchor=west] (winC) at (11.0,4.65)
  {near window\\8,192 tokens\\at true positions};
\node[cell, fill=gray!18, minimum width=2.0cm, minimum height=1.3cm, anchor=west] (req) at (14.5,4.65)
  {request,\\then the\\generated tail};
\draw[decorate, decoration={brace, amplitude=3pt}] (16.5,3.85) -- (0,3.85);
\node[lab, anchor=north] at (8.25,3.65) {trained range $L$: every cell is at a position inside it; the deepest slot, $8{,}192 + 16{,}384 + 129 + 2 = 24{,}707$, is inside \Mistral{}'s 32,768};
\end{tikzpicture}}
\caption{The life of a far unit, numbered in the order it happens. (a) Store: prefill runs the near window over the
document; the keys and values of every token that leaves the window go to the host exact, as written at the token's
true position, and a whole word above 15 nats of surprisal enters the word index with its sentence unit (1). The
document is unbounded and the host store grows with it. (b) Retrieve: before the request is encoded, its rare words and
the seed look up the index (2); the postings name the units, and the store supplies their keys and values. (c) Insert:
a unit is placed at a slot $s$ in the far window, just beyond the near window and inside the trained range, its keys
re-rotated from $p$ to $s$ plane by plane (3); the slow planes address it at $s$, and the order inside it is as
written. The request is then encoded over the far window. A rare word the model generates looks up the index the same way (4); the cache is then rewound to the start of the request, and the request and the generated tail are encoded again, exactly, with the fetched units present. The GPU cache holds about 25K cells per layer at any document length.}
\label{fig:design}
\end{figure}
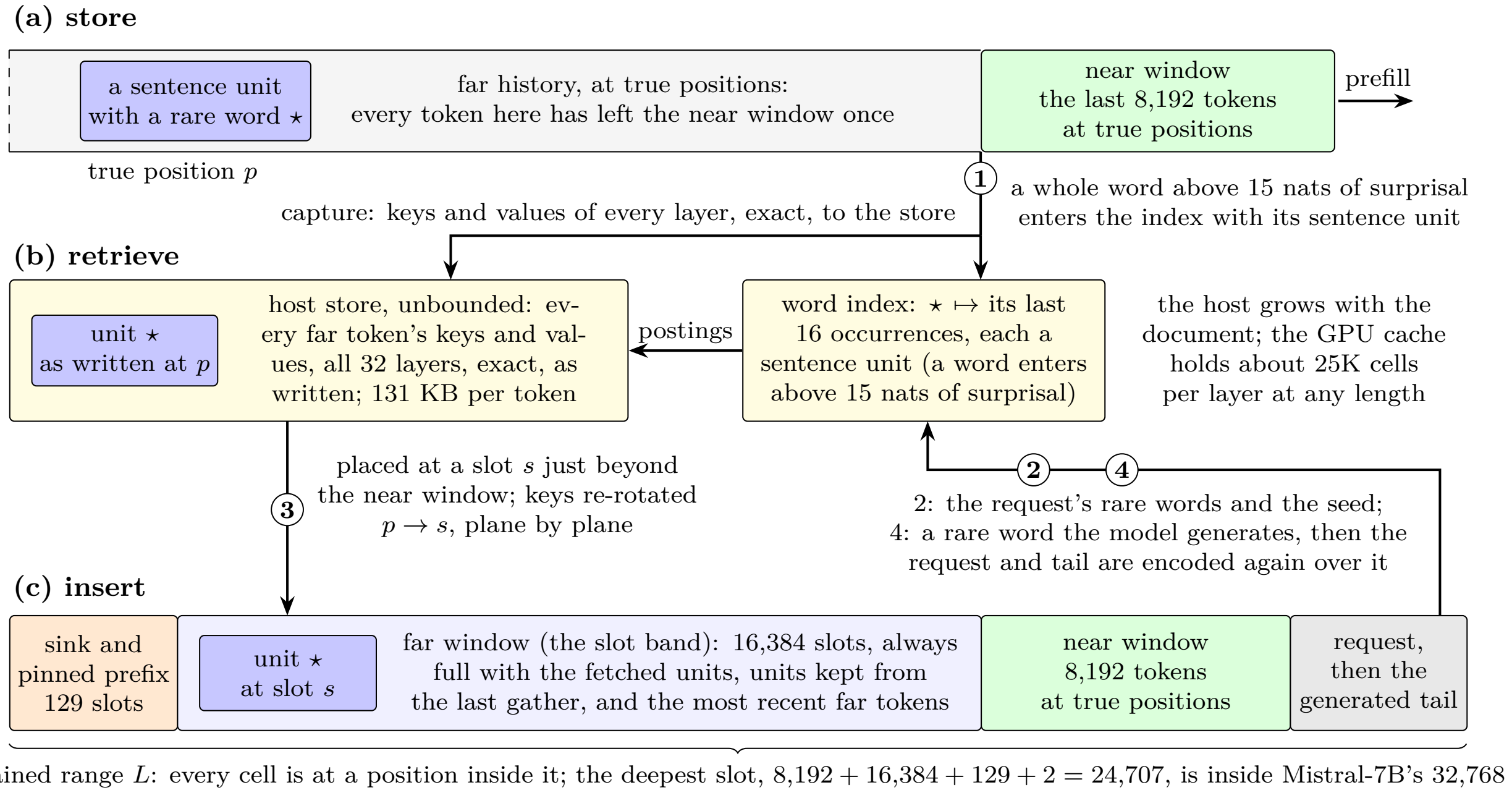

This paper describes how to do that. We split the window into two parts, near and far. The near window is the last
8,192 tokens, kept at their true positions as the model would normally see them. The far window is a fixed band of
16,384 slots inside the trained range whose contents we choose for each request. Every key and value that leaves the
near window is stored on the host exactly as written; the GPU cannot hold a million tokens of them, but host memory
can. We index this history by its unusual words, those whose surprisal under a running model of the document exceeds
a threshold, and keep each word's last 16 occurrences with the sentence around each. When a request arrives we look
up its unusual words, fill the far window with the sentences in which they last occurred, and only then encode the
request, so that the passages are present when the question is read. When the model generates an unusual word while
answering, we fetch that word's sentences as well, rewind to the start of the request, and encode the request and
the tokens generated so far again with the sentences present; the result is identical to what a fresh pass would
produce. Every setting is derived from the model's configuration. Figure~\ref{fig:design} follows one passage through
the store, the look-up and the far window. We implemented this as a small patch to \llamacpp{}~\citep{llamacpp}.\footnote{Our
patch is released as open source at \url{https://github.com/mark-oskin/unbounded-attention}.}

The engine reads planted needles, a seven-digit code hidden in a long document and asked for by name, at 99 to 100
of 100 from 32K to one million tokens, and when three codes of the same form are planted and one is asked for, at
88 to 100 of 100. On RULER~\citep{hsieh2024ruler}, a public long-context benchmark of thirteen tasks, it scores 70.8
against 69.9 for ordinary full attention on \Llama{} at 128K, and 67.8 against 75.5 on \Mistral{} at 32K. At 128K it
reads a document at 1.85 times the speed of unmodified \llamacpp{} and produces tokens at 1.65 to 1.68 times its
speed, because every query attends over about 25K cells rather than 128K, and a million tokens run on one H200 at
8,222 tokens per second of reading and 149 of generation. The limits are specific: a question whose answer needs
two hops is bounded on \Mistral{} by the first hop, which the near window alone loses, and a description alone does
not match at distance, as under full attention, unless the model first names the thing in the near window. The
technique generalizes beyond one document: the far window holds keys and values rather than text, so any text
encoded once through the model can enter the window later, which turns retrieval-augmented
generation~\citep{lewis2020rag,borgeaud2022retro,ram2023incontextralm} into retrieval into the attention rather than
into the prompt.

Our contributions, in the order the paper makes them, are these.
\begin{enumerate}
\item A census of what attention does, over every head of a 7B model reading long prose: most of its mass routes
within the nearest 64 tokens, a few heads in the middle of the stack do the far reading by copying from an earlier
occurrence of the current word, and the sink token is where a head puts its mass when it has nothing to say
(Section~\ref{sec:what}).
\item How a far read works in a stock rotary model: it is matched through the slow planes and tolerates any
position inside the trained range, it is bound at the moment the question's key word is encoded, it needs the
model's own values, and it is lexical, with reads by description ending at the front of the window
(Sections~\ref{sec:planes}--\ref{sec:descriptions}).
\item A serving design that keeps the recent window dense, stores every far key and value on the host, and fills a
bounded far window from a word index before each request, with its settings derived from the model's own
configuration, and its implementation as a small patch to \llamacpp{} (Sections~\ref{sec:design}
and~\ref{sec:implementation}).
\item Measurements of the implementation from 32K to one million tokens: planted needles and two-hop questions on
100 documents per cell, compared with ordinary dense attention and with a recent window alone; the 13 tasks of the
RULER long-context benchmark on both models; and throughput against unmodified \llamacpp{} (Section~\ref{sec:results}).
\item A way to reach far context by description rather than by word, in which the model first names the described
thing and the name fetches the far text, with the cost of a far window kept always full measured on it
(Section~\ref{sec:concept}).
\end{enumerate}
Section~\ref{sec:limitations} states the limitations of the measurements. Section~\ref{sec:discussion} discusses how the
approach can improve RAG by pre-encoding retrieved text through the model, so that it enters attention as keys and
values rather than as prompt. Section~\ref{sec:related} examines the extensive related work in this space, and we
conclude in Section~\ref{sec:conclusion}.

\secbarrier

\section{How attention behaves}
\label{sec:geometry}

In this section we explore how a stock transformer uses attention: where its attention goes, which reads carry information, and what a read needs from the key it lands on. We collect
the data in this section from two stock models: \Mistral{}~\citep{jiang2023mistral}, trained to 32K,
and \Llama{}~\citep{grattafiori2024llama3}, trained to 128K, with \LlamaBig{} where model size is the question.
We fine-tune nothing and change no weight. We measure with two code bases: a PyTorch harness with full control
of the cache and its masks, and a modified \llamacpp{} with the cached positions editable. Several of our measurements present a
far unit to the model at a fixed position just past the near window instead of at its true distance, called
\emph{placed in range} below; Section~\ref{sec:placement} shows that the placement changes nothing about what is
read. The section opens with an accounting of what attention spends its mass on (Section~\ref{sec:what}) and then
measures what the planes do (Section~\ref{sec:planes}), where a far key may sit (Section~\ref{sec:placement}), when
it must be present (Section~\ref{sec:residency}), what its value must be (Section~\ref{sec:values}) and which reads
survive at range (Section~\ref{sec:descriptions}). Section~\ref{sec:implications} summarizes and points the way forward
for how to exploit attention's behavior and limitations.

\subsection{What is attention doing?}
\label{sec:what}

\begin{figure}[htbp]
\centering
\includegraphics[width=\textwidth]{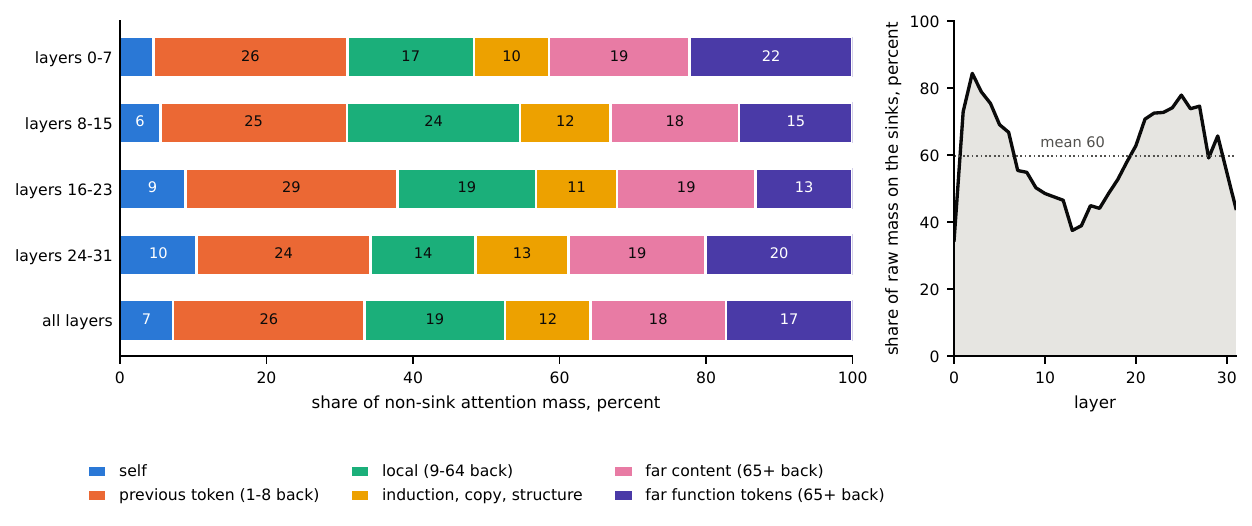}
\caption{Left: the six families of Table~\ref{tab:census} as shares of non-sink mass, by layer group and pooled. The
mix hardly moves with depth: routing (self, previous and local) holds 48--57~percent in every group, the far content
read 18--19~percent, far function tokens 13--22~percent. Right: the share of raw mass, sinks included, that lands on
the sink keys at each layer, 60~percent on average, 84~percent at layer 2 and 35--39~percent at layers 0 and 13.}
\label{fig:census}
\end{figure}

Through training, attention heads specialize: each learns input and output projections that make it perform a function. Before changing how they work, it is worth asking what they do now.
We took a census of all 1,024 heads of \Mistral{} over 100 windows of 2,048 tokens drawn from four
held-out Project Gutenberg novels, 102,400 queries in all; every share below is for windows of that length. For each
head and each query, we assign the head's mass to one of twelve operations by where the key sits relative to the
query: the query's own token, the previous token, each of the tokens two to eight back, the bands 9--16 and 17--64
back, an induction target (the token after the last occurrence of the current token, keyed on the token or on a
mid-layer state), a copy target (an earlier occurrence of the current token, or the token before it), the matching
quote or bracket, a sentence or paragraph start, a number or format token, and content tokens 65 or more back. Mass
on the sinks (the beginning-of-sequence token, the window's first 64 positions, newlines, and any key taking more
than 5~percent of the window's mass) is excluded from every share and counted separately as the head's rate of
abstention. Mass on function tokens 65 or more back matches none of the twelve and is the residual.
Table~\ref{tab:census} gives the pooled shares, Table~\ref{tab:heads} the heads and the sinks behind them, and
Figure~\ref{fig:census} shows how the shares move with depth. ``Far'' in the census means 65 or more tokens back
inside a 2K window; ``at range'' in the rest of the paper means beyond the 8K near window. The link between the two
is by layer rather than by distance: the layers whose heads read far content inside the window are the layers where
a planted item is read at range, as the lookup paragraph below shows; head-level overlap between the two was not
measured.

Every head needs a \emph{sink}.  This is a token that is (nearly) always in the history that the head can focus on when
it is not going to return a result.  During the softmax, if no other result is worth carrying into the residual the sink is chosen, and its value is innocuous.  Without the sink the head injects useless noise into the residual and that noise
degrades the prediction accuracy.  Sixty percent of the raw mass lands on the sink keys, most of all in the early layers (Figure~\ref{fig:census},
right; Table~\ref{tab:heads}). The sinks have enormous residual
norms~\citep{xiao2024streamingllm,sun2024massiveactivations}: at layer 2 of \Mistral{} the beginning-of-sequence
token's residual and the first newline's are hundreds of times the median token's, while later newlines are
ordinary. The mass they hold is not a read: spreading a head's sink mass uniformly over its last 64 keys costs more
than zeroing the head at the layers tested, so the sink sets the softmax floor, and a head that puts its
mass there has chosen to say nothing.  Besides the sink, the first few dozen tokens of the document serve as \emph{anchors}: keys that carry no content the query needs but hold the softmax's denominator where the model expects it, and the far function tokens of Table~\ref{tab:census} play the same role. The anchor is less specific than the sink, and the first few dozen tokens suffice.

\begin{table}[htbp]
\centering\footnotesize
\caption{Where non-sink attention mass goes, pooled over all 1,024 heads of \Mistral{} on 102,400 queries from 100
windows of 2,048 tokens. Each head's mass on each query is assigned to the operation whose key it lands on; sinks are
excluded from every share. The twelve named operations cover 82.7~percent of the non-sink mass. Function tokens 65 or
more tokens back match none of them and hold the remaining 17.3~percent.}
\label{tab:census}
\begin{tabular}{@{}llr@{\hspace{1.2em}}llr@{}}
\toprule
family & operation & share & family & operation & share \\
\midrule
routing   & self (the query's own token)     & 7.2\%  & lookup    & far content, 65 or more back         & \textbf{18.4\%} \\
routing   & previous token                   & 8.1\%  & anchoring & far function tokens, 65 or more back & 17.3\% \\
routing   & tokens 2--8 back                 & 18.1\% & induction & token after the last occurrence      & 1.7\% \\
routing   & tokens 9--16 back                & 6.4\%  & induction & same, keyed on a mid-layer state     & 1.7\% \\
routing   & tokens 17--64 back               & 12.7\% & copy      & an earlier occurrence of the token   & 1.5\% \\
structure & sentence or paragraph start      & 2.5\%  & copy      & the token before that occurrence     & 1.0\% \\
structure & number and format tokens         & 3.2\%  & structure & the matching quote or bracket        & 0.1\% \\
\bottomrule
\end{tabular}

\end{table}

\begin{table}[htbp]
\centering\footnotesize
\caption{The heads and the sinks behind the census. The same 1,024 heads and 102,400 queries as Table~\ref{tab:census}.
A head's best operation is the one that takes the largest share of its non-sink mass.}
\label{tab:heads}
\begin{tabular}{@{}>{\raggedright\arraybackslash}p{5.4cm}>{\raggedright\arraybackslash}p{10.6cm}@{}}
\toprule
measure & result \\
\midrule
\multicolumn{2}{@{}l}{\itshape the heads (all 1,024 of \Mistral{})} \\
purity, by the share of a head's mass on its best operation & 1 near-pure, 18 mixed, 1,005 diffuse (residual excluded); 1, 91 and 932 with far function tokens counted as an operation; the median head's best operation holds 23~percent of its non-sink mass \\
best operation, heads ranked & far content 490; the previous token 214; the band 17--64 back 208; the query's own token 69; the other six operations 43 together \\
the long tail of mass across heads & 323 heads carry half of the non-sink mass, 619 carry 80~percent, 860 carry 95~percent \\
the twelve operations folded into four families & 191 heads near-pure local, 159 mixing two families, 674 mixing local reading with far content and far function tokens \\
heads with a lift of at least 3 on a single-token aligned match (Table~\ref{tab:prefix}) & 107; of them 101 scale with the match length by at least 1.5 times, five are flat and one falls; the duplicate alignment gives the same picture (pooled 1.72, 3.80 and 6.08 at $m = 1$, 2 and 3; the heads that fall there are layer-0 heads); all 13 shared reader heads scale \\
\midrule
\multicolumn{2}{@{}l}{\itshape the sinks} \\
share of raw mass on the sink keys & 0.598 over the 102,400 queries; 84~percent at layer 2, 35--39~percent at layers 0 and 13 \\
rate of abstention per head (at least 90~percent of the mass on the sinks) & quartiles 0.4, 3.8 and 22.6~percent \\
residual norm of the sink tokens at layer 2, against the median token's & the beginning-of-sequence token 614 times, the first newline 712 times; the second, third and tenth newlines and the median over 502 newlines about one times \\
spreading a head's sink mass uniformly over its last 64 keys & costs more than zeroing the head at layers 0 and 8 \\
\bottomrule
\end{tabular}

\end{table}

Routing that builds the current position's state, the query's own token, the previous eight tokens and the two
local bands, is half of the non-sink mass. The far content read is under a fifth, far function tokens about the
same, and induction, copy and the structural operations together about a tenth. The heads carrying that mass are
mixtures rather than instances of one operation (Table~\ref{tab:heads}): by the share of a head's mass on its best
operation, almost every head is diffuse, and the median head's best operation holds under a quarter of its non-sink
mass. Ranked by best operation, nearly half of the heads read far content first and most of the rest route locally.
The mass has a long tail across heads as well: a third of the heads carry half of the non-sink mass. A head spreads
its mass smoothly over distance, so the near-purity that head-labeling studies report~\citep{voita2019heads,clark2019bert}
appears here only when the local operations are merged into families. Every head runs several operations at once,
and the softmax decides which one is heard.

The self, previous-token and local operations move information into the current position.
They do not forecast, per se, but rather act as functions that gather information that might be used (Table~\ref{tab:rules}). For each of the five biggest operations, the 16 heads carrying
most of its mass were turned into a rule in two forms, ``copy what followed the current token inside the attended
span'' and ``the token after the attended span,'' and scored by the top-1 agreement with the model the rule adds to a
fixed lexical forecaster on held-out positions, with the heads chosen on the other half of the novels. The self, previous-token
and near-band operations add almost nothing in either form, and the 17--64 band adds less than the best control that
uses no attention. Every operation in this family reads within 64 tokens of the query.

\begin{figure}[htbp]
\centering
\includegraphics[width=0.78\textwidth]{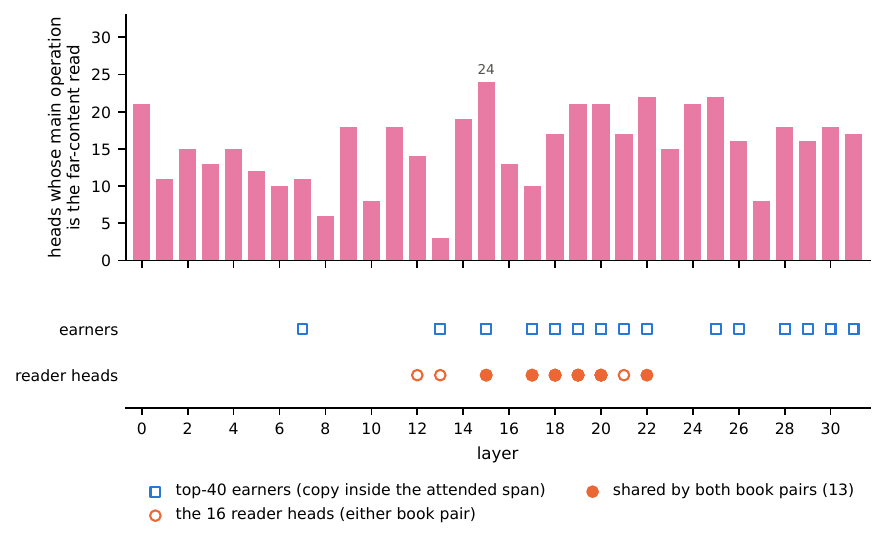}
\caption{Where the far readers live. Top: the number of heads in each layer whose best operation is the far content
read, 490 of 1,024 in all; every layer has some, 145 of them in layers 15--22. Bottom: the 16 reader heads chosen on either half of the novels (open circles) and the 13 chosen on both (filled), the union in layers 12--22 and the shared 13
in layers 15--22; and the 40 heads whose attended span forecasts the next token best (squares), every one of which
forecasts by copying what followed the current token inside its span.}
\label{fig:readers}
\end{figure}

\begin{figure}[htbp]
\centering
\includegraphics[width=0.9\textwidth]{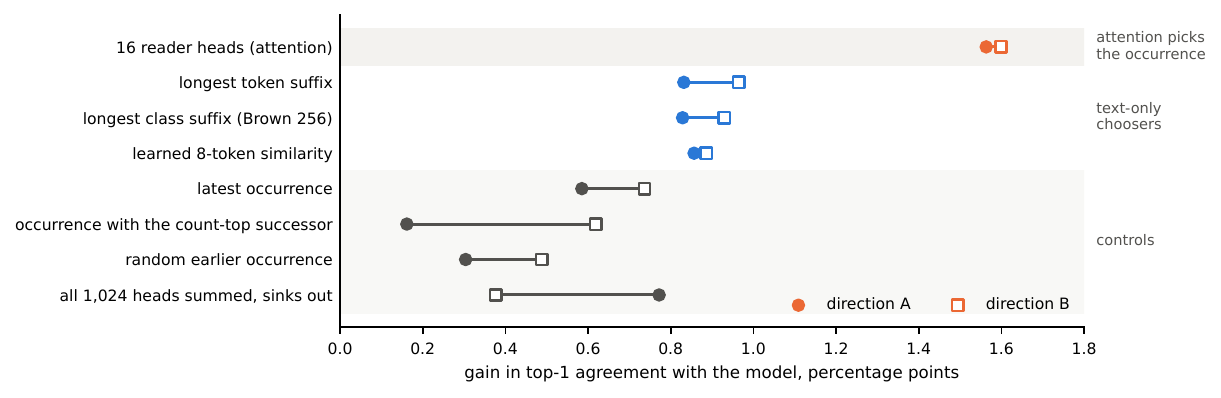}
\caption{What the reader heads' choice of occurrence is worth. Every row copies what followed the current token at
one earlier occurrence of it; the rows differ only in who chooses the occurrence. Gain is the top-1 agreement with the
model added to a fixed lexical forecaster on the held-out half of the novels (direction A: 237,568 positions; direction B: 230,633), choices and thresholds fixed on the other half. The 16 reader heads add $+1.56$ / $+1.60$ points; the best
text-only chooser $+0.96$; the latest occurrence $+0.59$ / $+0.74$.}
\label{fig:chooser}
\end{figure}

The far content read is the one operation whose attended span forecasts the next token beyond what a count of earlier occurrences already gives.
We chose sixteen reader heads, copy heads in the plain sense that each reads from an earlier occurrence of the current token, on one half of the four novels and tested them on the other half, in both directions; 13 of the 16 are chosen on both halves, all in layers 15 to 22 (Figure~\ref{fig:readers}). The rule ``copy what followed the current
token at the occurrence these heads attend to'' fires at a tenth to a fifth of positions and adds more top-1
agreement than any rule built without attention (Table~\ref{tab:rules}). Figure~\ref{fig:chooser} sets it against
the same copy with the occurrence chosen without attention: the latest occurrence, the occurrence with the most
frequent successor, a random occurrence, and all heads summed each recover part of the gain; four text-only choosers
of the occurrence land within a tenth of a point of each other at a little over half of what the heads give, and the
heads still add most of their gain on top of the best of them. These are the retrieval heads of
\citet{wu2024retrievalhead} seen from the forecasting side. The same exhaustive search over all heads finds that
every one of the best-forecasting heads does so through this copy, never through the token after the span, and that
they sit in the middle and late layers. A planted item read at range lands on the same layers: a working read of a
far needle in the harness puts its attention on the far keys with a peak at layer 15.

\begin{table}[htbp]
\centering\scriptsize
\caption{What each operation forecasts. Every rule copies from the span the named heads attend to, or copies at an
occurrence chosen without attention; the gain is top-1 agreement with the model added to a fixed lexical forecaster
on the held-out half of the novels, in both directions, with heads and thresholds chosen on the other half.}
\label{tab:rules}
\setlength{\tabcolsep}{4pt}
\begin{tabular}{@{}>{\raggedright\arraybackslash}p{5.2cm}>{\raggedright\arraybackslash}p{4.2cm}>{\raggedright\arraybackslash}p{6.4cm}@{}}
\toprule
rule & form & result (direction A / B) \\
\midrule
\multicolumn{3}{@{}l}{\itshape rules built from the 16 heads carrying most of each operation, scored on about 470,000 held-out positions} \\
self, the previous token, 9--16 back & copy what followed the current token inside the span; the token after the span & at most $+0.08$ points in either form \\
17--64 back & the same two forms & $+0.29$ / $+0.27$ against $+0.77$ / $+0.74$ for the best control that uses no attention; ``the token after the span'' at precision about 0.01 for every local operation \\
far function tokens, 65 or more back & the 16 heads carrying it most (layer-0 heads L0H25, L0H23, L0H22 and L0H19 lead) & $+0.14$ / $+0.06$, below every no-attention control \\
far content: the 16 reader heads & copy what followed the current token at the occurrence the heads attend to & fires at 20 / 10~percent of positions, precision 0.47 / 0.62, $+1.56$ / $+1.60$ points \\
\midrule
\multicolumn{3}{@{}l}{\itshape the same copy with the occurrence chosen without attention (Figure~\ref{fig:chooser})} \\
the latest occurrence & & $+0.59$ / $+0.74$ \\
the occurrence with the most frequent successor & & $+0.16$ / $+0.62$ \\
a random occurrence & & $+0.30$ / $+0.49$ \\
all 1,024 heads summed, sinks removed & & $+0.77$ / $+0.38$ \\
four text-only choosers: the longest matching token suffix, the longest suffix over 256 word classes, a learned similarity over the previous eight tokens, a recency tie-break & & $+0.83$ to $+0.96$, within 0.1 point of each other, 53--60~percent of what the heads give; the heads add $+0.90$ / $+0.94$ on top of the best of them \\
\midrule
\multicolumn{3}{@{}l}{\itshape where the heads are right and the longest-suffix chooser is wrong} \\
share of the rule's firings & & 9.7 / 10.8~percent (4,531 / 2,614 cases) \\
preceding tokens the chosen occurrence shares with the query & & none in 78 / 80~percent of cases, at most one in 91 / 92~percent \\
the chosen occurrence is the latest one & & 40 / 35~percent; over all firings the heads pick the chooser's own occurrence 44 / 51~percent of the time \\
distance of the chosen occurrence & 10th / 50th / 90th percentile & 13 / 72 / 695 and 17 / 127 / 870 tokens \\
\midrule
\multicolumn{3}{@{}l}{\itshape the best-forecasting heads, exhaustive over all 1,024, and the read at range} \\
the 40 heads whose attended span forecasts best & & every one through this copy, never through the token after the span; layers 7 to 31, 37 of the 40 in layers 15--22 and 25--31 \\
a working read of a far needle at 128K (harness, $n = 16$) & attention on the far keys & mean 0.067, peak 0.235 at layer 15 \\
\bottomrule
\end{tabular}

\end{table}

The reader heads do not simply match the last few tokens before the query. A text-only rule that picks the earlier
occurrence with the longest matching suffix, the same tokens before the occurrence as before the query, picks the
same occurrence as the heads about half the time. Where the two differ and the heads are right, about a tenth of the
cases, the occurrence the heads copy from usually shares no preceding token with the query at all, and it is usually
not the most recent occurrence (Table~\ref{tab:rules}). It is typically the same scene or the same referent, tens to
hundreds of tokens back: ``to pay the'' is completed from ``to tell the [coach]'' in the chariot scene just before,
and ``said the'' from ``said my convict to the [sergeant]'' hundreds of tokens back. The heads choose by content that
a suffix match does not see. The claim has limits. The cases where the two differ are, by construction, cases of low
suffix overlap, and the base rate of such occurrences among all candidates was not measured, so the heads choose
beyond suffix matching rather than apart from it. The comparison also considers only earlier occurrences of the
current token; that the far read is lexical at all rests on the verbatim-against-description rows at the end of this
section and in Section~\ref{sec:descriptions}.
Section~\ref{sec:placement} shows that the read is position-free as well.

Induction, keyed on the token or on a mid-layer state, copy of an earlier occurrence and of its predecessor, and the structural operations (quotes and brackets, sentence and paragraph starts, number and format tokens) hold about
a tenth of the non-sink mass between them.  \citet{feucht2025dualroute} separate, by
ablation, a token route that copies verbatim from a concept route that carries meaning across paraphrase and
translation. The census is consistent with that account from the outside, as a behavioral dissociation: the token
route is a count, which token followed the current token before and how often, and a count of earlier successors
built from the window alone reproduces it, which is why the latest occurrence with no attention recovers a good part
of the reader heads' gain and why the four text-only choosers cannot be told apart. What the reader heads add is a
choice a count cannot make. No head was ablated here, so the identification stays behavioral. The classic induction
head~\citep{olsson2022induction,elhage2021framework} is the token route seen one head at a time; at under two percent
of the mass it is a small part of what attention does on prose.

The induction route is longer than one token. For every query $x$ and earlier key $p$ in the census windows, let $m$
be the length of the aligned match that ends at the key's predecessor, the number of consecutive $k \geq 0$ with
$\mathrm{tok}(x - k) = \mathrm{tok}(p - 1 - k)$, so that $m = 1$ is the classic induction target and $m = 3$ a key
that follows a three-token repeat of the query's own recent past. Table~\ref{tab:prefix} gives the attention on such
keys as a lift over the head's mean attention at the same distance, with the sinks excluded. Pooled over all heads
and weighted by mass, a key that follows a two-token match draws more than twice the attention of one that follows a
single-token match, and a three-token match more again. Nearly every head with a strong single-token lift scales with
the match length, and the duplicate alignment ($\mathrm{tok}(x - k) = \mathrm{tok}(p - k)$) gives the same picture
(Table~\ref{tab:heads}). All 13 shared reader heads scale, some by an order of magnitude from a one-token match to a
two-token match. \citet{dangelo2026ngram} find the same exponential weighting of a context match by its overlap in
induction circuits trained on synthetic data, and the generalized induction head of \citet{kim2025gim} adds fuzzy
matching to the count; the census shows the same rule at work on a 7B model reading prose. The reader heads therefore
do two things that the chooser analysis above keeps apart. When the query's recent past repeats somewhere earlier,
they weight the longer repeat more, which is induction over $n$-grams rather than over one token, and when no such
repeat exists, the case in four of five of their disagreements with the suffix chooser, they still choose an
occurrence, by content the suffix does not see.

\begin{table}[htbp]
\centering\scriptsize
\caption{Attention against the length $m$ of the aligned match, \Mistral{} on the census windows (100 windows of 2,048
tokens, 192 million query-key pairs after the sink exclusions). Lift is the attention on a key over the head's mean
attention at the same distance bin, so 1 is no preference; restricting to keys 65 or more tokens back changes no
entry by more than a point. The pooled row weights every head by its mass; the 13 rows are the shared reader heads of
Figure~\ref{fig:readers}.}
\label{tab:prefix}
\begin{tabular}{@{}l@{\hspace{1.2em}}rrr@{\hspace{1.6em}}rrr@{\hspace{1.6em}}c@{}}
\toprule
 & \multicolumn{3}{c}{induction alignment, lift at $m =$} & \multicolumn{3}{c}{duplicate alignment, lift at $m =$} & mass on $m \geq 1$ \\
\cmidrule(lr){2-4}\cmidrule(lr){5-7}
head & 1 & 2 & $\geq 3$ & 1 & 2 & $\geq 3$ & ind.\ / dup. \\
\midrule
all 1,024 heads, mass-weighted & 1.71 & 3.80 & 4.6 & 1.72 & 3.80 & 5.6 & \\
\midrule
L15H21 & 2.1 & 7.9 & 7.5 & 16.3 & 100.0 & 124.3 & 0.03 / 0.29 \\
L17H26 & 5.1 & 19.8 & 19.0 & 1.8 & 5.5 & 5.7 & 0.08 / 0.02 \\
L17H27 & 1.9 & 6.1 & 6.4 & 5.5 & 33.0 & 40.9 & 0.03 / 0.09 \\
L18H0 & 20.7 & 176.3 & 147.0 & 0.6 & 5.6 & 5.9 & 0.51 / 0.01 \\
L18H2 & 15.9 & 79.7 & 86.1 & 0.7 & 3.3 & 4.2 & 0.31 / 0.01 \\
L18H11 & 2.2 & 4.9 & 4.9 & 1.6 & 4.4 & 6.0 & 0.03 / 0.02 \\
L19H8 & 5.3 & 20.5 & 22.7 & 1.4 & 3.6 & 4.4 & 0.09 / 0.02 \\
L19H17 & 4.8 & 11.7 & 12.0 & 1.1 & 3.5 & 4.1 & 0.06 / 0.02 \\
L19H19 & 2.0 & 5.9 & 7.1 & 1.7 & 4.9 & 6.0 & 0.03 / 0.02 \\
L20H6 & 7.0 & 52.3 & 66.7 & 1.1 & 4.9 & 5.9 & 0.16 / 0.02 \\
L20H20 & 2.0 & 4.3 & 4.2 & 3.6 & 9.2 & 12.7 & 0.03 / 0.05 \\
L20H23 & 1.9 & 4.3 & 3.9 & 11.1 & 47.4 & 64.1 & 0.02 / 0.18 \\
L22H29 & 3.2 & 7.8 & 8.9 & 0.9 & 2.0 & 2.7 & 0.04 / 0.01 \\
\bottomrule
\end{tabular}

\end{table}

Function tokens 65 or more tokens back, commas, articles and newlines far from the query, hold about a sixth of the non-sink mass but match no operation. The heads that carry most of it are layer-0 heads, and as a rule through the
16 heads that carry it most, that mass adds less than every no-attention control (Table~\ref{tab:rules}). It behaves
as the sink does: mass parked on keys that carry no content the query needs, holding the softmax's denominator where
the model expects it. An anchor the model cannot read is worse than none: outside the trained range the sink itself
stops holding the floor (Section~\ref{sec:placement}).  In other words, the grammatical scaffolding a query uses comes from within 64 tokens; far function tokens hold mass and carry nothing the query reads.

We measured everything above on prose. A planted item tells the same story at range
(Table~\ref{tab:descriptions}). A question that names the item's handle word verbatim is answered about nine times
in ten 32K back on \Mistral{} and 128K back on \LlamaBig{}. A question that describes the same item with no shared content word is answered nine times in ten or more when the item is a thousand tokens back and almost never 32K and 128K back in
every model measured, each with three same-format decoys present and full attention over the whole document
(Section~\ref{sec:descriptions}).  In other words, a question about ``the pet'' does not reach a far passage about ``the cat'', though a cat is a pet; only the word ``cat'' does. What reaches far back in the attention window is the exact word rather than its description.

\subsection{The planes}
\label{sec:planes}

Rotary attention~\citep{su2021roformer} groups dimensions into pairs. Each pair of key dimensions is a \emph{plane}
that turns with position at its own rate: on \Mistral{}, with a rotary base of $10^6$ and 64 planes per head, the
fastest plane completes a turn every six tokens and the slowest every five million. A key may be viewed as 64 clock hands. The slow hands say which neighborhood of the document the key belongs to; the fast hands say which seat
inside it.

A query and a key agree on a plane only while the plane's turn between them is under about a radian, so the number
of planes a key is coherent on falls with its distance, from most of them a few hundred tokens back to a quarter at
the edge of the 32K trained range (Table~\ref{tab:planefacts}). The count follows from the rotary schedule alone.
The slow planes serve as the index card: they carry the match between a query and a far key, and they carry it at any
distance. The fast planes serve as the page: they turn once in a few tokens, and whatever order a span's tokens carry is
already in the span's keys and values by the time the span leaves the window. The measurements below separate the
two jobs in time: the slow planes do their addressing when a query reads, and the order inside a span is settled
earlier, while the model writes the span's keys.

The 15 slowest of \Mistral{}'s 64 planes carry most of the score between a query and a far key, two and a half times
what they would carry if the planes were interchangeable, and the share is the same at every distance.
Position-sensitive heads, those whose far scores change under a small jitter of the key positions, concentrate in
layer 1, and more than a third of the heads are flat under the jitter. On 8K documents under dense attention at
true positions, keeping only the 15 slowest planes of every key older than 1,024 tokens costs a few hundredths of a
nat, less when the cut starts later or keeps more planes, and planted needles read in full under it; keeping the 15
\emph{fastest} planes instead costs several nats and reads nothing, because it destroys the sink
(Table~\ref{tab:planefacts}). The far match lives in the slow planes, which is the observation behind analytic
plane cuts for far keys~\citep{hayashi2026atflash,barbero2024roundandround,hong2024tokendistance}, the finding of
\citet{chiang2025dimension} that long-distance retrieval rides on the lowest-frequency dimensions, and the
key-channel pruning family~\citep{zhang2025leank,wang2026prism,yao2026thinkeys,xu2025think}.

The fast planes carry signal out to a few hundred tokens, and that signal is consumed while the span is near: the
cut above costs less the later it starts, so what the fast planes contribute is concentrated in the near past. Who
reads them there is visible in the weights: most key-carrier reader heads have their strongest upstream writer in
layers 0--1, and those writers attend one token back far more than the head population does, previous-token heads
found from the weights alone. What a far key then carries of its neighborhood is an ordered bag: the contributions
of the tokens at small offsets enter the key at falling norms, and a question about the order of two far mentions
is answered well above chance with every far position collapsed onto one. The order of the tokens inside a span is
written into the span's own keys and values while the span is inside the window; the weights name the local
previous-token heads of Section~\ref{sec:what} as its writers, by inference from the writer-reader graph rather than
by a write-time test.

Every far cell, the needle's seven digits included, placed on one identical position reads every single needle
(Table~\ref{tab:planefacts}). A far key cut to its 22 slowest planes still reads three-key
discrimination in full (Table~\ref{tab:planes}). And rotating the fastest planes of every far key by the token's
offset inside its own span rather than by the key's position reads every single needle and every three-key
discrimination at the same loss as rotation by position. The one thing a far span cannot do without is the
distinctness its tokens were written with: far keys synthesized from one key per word, so that every digit of a
code shares one key, are addressed correctly and read nothing. Slow planes address at read; order is settled at
write.

\begin{table}[htbp]
\centering\scriptsize
\caption{What the planes carry. \Mistral{} unless stated; the plane-cut rows are 8K documents under dense attention
at true positions, the agreement rows the harness at 32K with $n = 16$ except where the table says \llamacpp{}.}
\label{tab:planefacts}
\begin{tabular}{@{}>{\raggedright\arraybackslash}p{5.6cm}>{\raggedright\arraybackslash}p{10.4cm}@{}}
\toprule
measurement & result (\Mistral{} unless stated) \\
\midrule
\multicolumn{2}{@{}l}{\itshape the slow planes address} \\
planes a key is coherent on, by distance (the plane's turn under about a radian) & 39 of 64 at 200 tokens back, 26 at 3,000, 21 at 10,000, 18 at 20,000 and 15 at the edge of the 32K trained range; from the rotary schedule alone \\
share of the query-key score carried by the 15 slowest planes & 59~percent, against the 23~percent that 15 of 64 would carry if the planes were interchangeable; the same at every distance (one 8K document, far keys 1K to 8K back) \\
heads flat under a $\pm 64$-token jitter of the far key positions & 391 of 1,024; the position-sensitive heads concentrate in layer 1 \\
keeping only the 15 slowest planes of every key older than 1,024 tokens (8K documents, dense attention at true positions) & $+0.032$ nats ($[+0.014, +0.051]$ over seven documents); $+0.007$ when the cut starts at 4,096; $+0.006$ and $+0.0007$ for 24 and 32 planes past 1,024; $+0.26$ for 8; the model-card rule that keeps every plane coherent at 1,024 tokens (31 planes) $+0.0014$; planted needles \rd{38}{38} under the 15-plane cut against \rd{100}{100} unpatched \\
keeping the 15 fastest planes instead & $+6.85$ nats and \rd{0}{30}: the sink's share of mass falls from 0.459 to 0.0005 \\
\midrule
\multicolumn{2}{@{}l}{\itshape the fast planes write the order} \\
who writes the fast-plane signal (from the weights) & of 248 key-carrier reader heads, 167 have their strongest upstream writer in layers 0--1; those writers' measured one-back attention is 0.49 against 0.018 for the head population \\
what a far key carries of its neighborhood & the tokens at offsets 0--4 enter the key at norms 4.9, 2.6, 1.8, 1.5 and 1.3 \\
the order of two far mentions, every far position collapsed onto one & answered at 0.750 $[0.598, 0.858]$, $n = 40$ \\
\midrule
\multicolumn{2}{@{}l}{\itshape order is settled at write} \\
every far cell, the needle's seven digits included, placed on one identical position & \rd{7500}{7500} single needles (Table~\ref{tab:placement}) \\
a far key cut to its 22 slowest planes & three-key discrimination \rd{16}{16} (Table~\ref{tab:planes}) \\
the 21 or 27 fastest planes of every far key rotated by the token's offset inside its own span rather than by position (32K, harness, $n = 16$) & \rd{16}{16} single and \rd{16}{16} three-key; loss 2.1940 / 2.1943 and 2.1938 / 2.1939 against 2.1942 for rotation by position \\
far keys synthesized from one key per word, so that every digit of a code shares one key & addressed correctly (0.7--0.99 of the far mass lands on the unit) and read at \rd{0}{16} \\
\bottomrule
\end{tabular}

\end{table}

Table~\ref{tab:planes} keeps the $k$ slowest planes of every far key and drops the rest, at four-bit keys (groups of
32, per-head centering) and four-bit values, in the harness on \Mistral{} at 32K with $n = 16$ documents, each holding
one seven-digit needle and two same-format competitors. Three-key discrimination holds in full from 64 planes down
to 22 and loses two reads at 15; single needles read in full at every rung; loss is identical to four decimals at
every rung. A far key cut to its 22 slowest planes still reads; cut to 15 it does not, and the cliff between the two
is shallow. The 22 planes are exactly those whose wavelength is at least $2\pi \times 8{,}192$ tokens, every plane
still coherent across a span the length of the near window. Bit width is free at every rung: four-bit keys and values
read the same single and three-key rows as eight-bit and as bf16. Loss does not resolve the cliff: across the whole ladder
it moves by less than a thousandth of a nat while the three-key read is what moves. The ladder is measured on
\Mistral{}; \Llama{}'s rungs are unmeasured under this protocol.

\begin{table}[htbp]
\centering\footnotesize
\caption{The plane ladder. \Mistral{} at 32K in the harness, $n = 16$ documents, one seven-digit needle and two
same-format competitors per document, the far unit placed in range and visible from the first token of the question;
keys kept in their $k$ slowest planes at four bits (groups of 32, per-head centering), values at four bits. Bytes are
the counted size of one token's keys and values at each rung; the 64-plane row is the full bf16 key and value.}
\label{tab:planes}
\begin{tabular}{@{}rrrrr@{}}
\toprule
planes kept (of 64) & bytes per token & single needle & three-key & loss, nats \\
\midrule
64 (bf16, uncut) & 131 KB & \rd{16}{16} & \rd{16}{16} & 2.1944--2.1948 \\
40 & 33.0 KB & \rd{16}{16} & \rd{16}{16} & 2.1944--2.1948 \\
32 & 30.0 KB & \rd{16}{16} & \rd{16}{16} & 2.1944--2.1948 \\
28 & 29.0 KB & \rd{16}{16} & \rd{16}{16} & 2.1944--2.1948 \\
22 (every plane coherent across 8,192 tokens) & 27.5 KB & \rd{16}{16} & \rd{16}{16} & 2.1944--2.1948 \\
15 & 24.8 KB & \rd{16}{16} & \rd{14}{16} & 2.1944--2.1948 \\
\bottomrule
\end{tabular}

\end{table}

\subsection{The far match is position-free}
\label{sec:placement}

The far match does not use the key's position (Table~\ref{tab:placement}). On \Mistral{} with read-time edits of the
cached positions, single planted needles read in full under 25 different edits, including every far key moved to one
position and the needle moved to the far edge of the trained range; moved beyond the range the read degrades, a
little at twice the range and badly at three times it. The one placement that fails is distance zero, the current
token's own phase. Presenting every far key at one position just beyond the near window, regardless of where it
really was, is therefore free: the whole far history placed that way reproduces full attention at 32K on loss and on
needles, where the same keys scored at their true graded distance lose half the needles, and a set of far units
placed together at one distance reads nearly all. Position-free storage with re-rotation at a new position is the
primitive of CachedAttention~\citep{gao2024cachedattention}, ReAttention~\citep{liu2025reattention},
MiniPIC~\citep{ordonez2026minipic}, LazyAttention~\citep{xia2026lazyattention} and
CacheFocus~\citep{lee2025cachefocus}; what is specific here is the measurement that the far match tolerates any
placement inside the range, degrades beyond about twice it and loses its anchors outside it, and that the keys
placed this way are read in the same layers as the reader heads of Section~\ref{sec:what}.

Moving a key is likewise free. A key rotated at one position and then re-rotated to another by the difference is the
same key as one rotated once at the new position: a single shift of the whole sequence by up to a million positions
changes the top-1 token at almost no decoded position, and the dense logits at that offset are the same to five
decimals (Table~\ref{tab:placement}). Inside the range, where a key sits is a number the model does not check;
Section~\ref{sec:impl:rerotation} checks the two ways an engine can make the move.

Outside the trained range the anchors fail, the sink first. A sliding window whose sink sits at its true distance
loses more than half a nat on \Mistral{} and more than a nat on \Llama{} by one million tokens; with the sink placed
in range the same window is flat to one million. The first 128 tokens of the document, kept as anchors at their true
distance, cost several nats at 1M. Inside the trained range a key may sit anywhere; beyond it the match degrades
with distance and the anchors the softmax stands on are lost.

\begin{table}[htbp]
\centering\footnotesize
\caption{Where a far key may sit. Read-time edits of the cached positions, placement of the whole far history, shifts
of the whole sequence, and the anchors outside the trained range.}
\label{tab:placement}
\begin{tabular}{@{}>{\raggedright\arraybackslash}p{5.6cm}>{\raggedright\arraybackslash}p{10.4cm}@{}}
\toprule
measurement & result \\
\midrule
single planted needles under read-time edits of the cached positions (\Mistral{}, \llamacpp{}, $n = 300$) & \rd{7500}{7500} under 25 different edits, including every far key moved to one position and the needle moved to 30K back; moved to 40K and 60K on the 32K-trained model 0.993 and 0.960; at 100K 0.213 (0.696 per digit); distance zero, the current token's own phase, is the one placement that fails \\
the whole far history placed at one position just beyond the near window (\Mistral{} at 32K, $n = 16$) & loss 2.1863 against 2.1857 under full attention; needles \rd{14}{16} against \rd{14}{16}; the same keys scored at their true graded distance 2.1937 and \rd{7}{16} \\
a set of far units placed together at one distance of 8,192 tokens & \rd{23}{24} \\
a single shift of the whole sequence by $2^{17}$ to $2^{20}$ positions & the top-1 token changes at no more than one of 256 decoded positions; key cosine at least 0.9992 \\
dense logits at a rotary offset of $2^{20}$ against offset 0 & differ by $6 \times 10^{-6}$ \\
a sliding window whose sink sits at its true distance, to one million tokens & $+0.58$ nats on \Mistral{} and $+1.15$ on \Llama{}; with the sink placed in range the same window is flat ($2.198 \pm 0.002$ and $2.4512 \pm 0.0005$ nats over six lengths from 32K) \\
the first 128 document tokens kept as anchors at their true distance, at 1M & $+3.9$ to $+5$ nats \\
\bottomrule
\end{tabular}

\end{table}

\subsection{Residency and binding}
\label{sec:residency}

When must a far unit be present? With a same-format competitor in the document, the answer is a moment rather than a
span: the unit must be visible to one particular query. Table~\ref{tab:residency} places the needle's unit in range
with no retrieval step, so that the unit is simply there, and masks it so that only queries from a chosen position
onward can see it, in two harnesses, with and without three same-format competitors in the document.

\begin{table}[htbp]
\centering\footnotesize
\caption{When the far unit must be visible. The unit is placed in range and masked to queries from the stated
position onward; the question names the handle word and asks for the seven-digit code. The needle harness is the
harness of the planted-needle rows above; the description harness is the one of Section~\ref{sec:descriptions}. With
competitors present the binding forms at the handle word's own query: one position earlier reads, one position later
reads nothing. The last group is the chain that no prefill regime binds.}
\label{tab:residency}
\begin{tabular}{@{}p{7.4cm}p{3.6cm}p{4.6cm}@{}}
\toprule
the unit is visible from & no competitors & three same-format competitors \\
\midrule
\multicolumn{3}{@{}l}{\emph{\Mistral{} at 32K, needle harness: $n = 16$ without competitors, $n = 8$ with}} \\
the handle word's own query & \rd{16}{16} & \rd{8}{8} \\
one / two / four tokens before the first digit, handle excluded & \rd{14}{16} / \rd{14}{16} / \rd{15}{16} & \rd{3}{8} / \rd{3}{8} / \rd{4}{8} \\
the query that predicts the first digit & \rd{8}{16} & \rd{2}{8} \\
6 / 20 / 256 tokens before the handle & \rd{16}{16} / \rd{16}{16} / \rd{16}{16} & \\
the window's tail only, evicted at the question & \rd{0}{16} & \\
throughout, evicted before the digits & \rd{0}{16} & \rd{0}{8} \\
\addlinespace
\multicolumn{3}{@{}l}{\emph{\Mistral{} at 32K, description harness: $n = 28$ without competitors, $n = 81$ with}} \\
one position before the handle & \rd{25}{28} & \rd{75}{81} \\
one position after the handle & \rd{0}{28} & \rd{0}{81} \\
the first token of the request / after the handle word, many units at once ($n = 56$) & & \rd{53}{56} / \rd{0}{56} \\
\addlinespace
\multicolumn{3}{@{}l}{\emph{\Llama{} at 128K, description harness: $n = 8$}} \\
one position before / after the handle & & \rd{7}{8} / \rd{0}{8} \\
\addlinespace
\multicolumn{3}{@{}l}{\emph{a chain reached only through a referring unit, the question sharing no rare word with the answer unit ($n = 16$)}} \\
encoded with full attention / the near window alone / only its last window, read at range & \multicolumn{2}{l}{\rd{0}{16} in every case, as the unreachable control} \\
the same chain inside the window & \multicolumn{2}{l}{\rd{16}{16}, four reads above a \rd{12}{16} format-copy floor} \\
\bottomrule
\end{tabular}

\end{table}

The binding forms at the handle's own query. With three competitors, a unit visible from the handle word is read in
full; visible from one, two or four tokens before the first digit, with the handle's query excluded, it is read about
half the time; visible only at the predicting query, a quarter of the time. Without competitors the requirement
relaxes but does not vanish: nearly every read survives a few tokens of lead, and half survive visibility at the
predicting query alone. The description harness draws the same line: with three competitors a unit visible from one
position before the handle is read nearly always on both models, and from one position after it never, with and
without competitors. Nothing earlier than the handle adds anything: visibility from 6, 20 or 256 tokens before it
reads in full each time, and residency in the window's tail alone, evicted at the question, reads nothing. The unit
must also stay: evicted before the digits, the read is lost in the harness and in \llamacpp{} alike, with the generated
name surviving in the residual stream and the verbatim copy lost. The line holds when many far units become visible
together: a set of units visible from the first token of the request reads nearly all, and the same set visible only
after the handle word reads none. This is the residency side of the query-aware contextualization result of
\citet{liu2024lostmiddle}, at fixed order and matched mass: the same content, the same position, only the moment of
visibility varied. It is also the mechanism behind the query-visibility audits of \citet{luo2026queryvis} and the
wait-to-attend memory of Attendre~\citep{yang2024attendre}, which keeps evicted queries so that a later key can still
be attended by them.

The binding also does not form at prefill time across units. In a chain whose question shares no rare word with the
answer unit and reaches it only through a referring unit, a document encoded with full attention, with the near
window alone, or with only its last window reads nothing at range in every case, as does the unreachable control,
against a full read for the same chain inside the window (Table~\ref{tab:residency}). The referring unit's cached
keys and values carry nothing of the answer unit under any prefill regime. A second hop is work the model does while
it generates.

The same rule governs a unit that becomes visible while the model is generating (Table~\ref{tab:events}). In
\llamacpp{}, with no unit supplied at the question and the needle's unit made visible only once the model has
generated its handle word, so that it first affects a prediction two tokens after the word closes, a plain copy
question reads the late arrival nearly always on both models with no further step. With one same-format worked
example in the request, neither model reads it at either length, with the unit present and unmasked. Encoding the request and the generated tail again with the unit present, which the engine does by rewinding the cache to the start of the request and rolling forward, recovers most of the reads on \Mistral{} and all of them on \Llama{}; the pass is identical in its logits to a fresh prefill on both models, and the design calls it the re-encode. It also takes the chained hop on \Mistral{} from about half to all. The description harness draws the same picture on the description side: a unit
that arrives at the model's own naming of it sits among the far keys, unmasked in all 32 layers, at a small fraction
of the attention a resident unit receives, and the verbatim read goes from almost none to all once the question is
re-encoded over it. Presence is not residency: under competition the binding forms when the handle's query runs,
and a unit that was absent then stays unread until that query runs again; without a competitor the late arrival is
read as it stands.

\begin{table}[htbp]
\centering\footnotesize
\caption{A unit that arrives mid-generation, with a same-format worked example in the request, is read reliably only
after the request and the generated tail are re-encoded over it. \llamacpp{}, with no unit supplied at the question: the needle's unit becomes visible only once the model has generated its handle word; $n =
16$ per cell, one run; a second run of the worked-example cells read \rd{12}{16}, \rd{15}{16},
\rd{15}{16} and \rd{16}{16}. The chained hop, in which the second name is given, $n = 7$ on \Mistral{} and $n = 5$ on
\Llama{}. The re-encode is identical in its logits to a fresh prefill on both models.}
\label{tab:events}
\begin{tabular}{@{}p{5.9cm}p{1.8cm}cccc@{}}
\toprule
 & & \multicolumn{2}{c}{\Mistral{}} & \multicolumn{2}{c}{\Llama{}} \\
\cmidrule(lr){3-4}\cmidrule(lr){5-6}
question & re-encode & 32K & 128K & 32K & 128K \\
\midrule
plain copy, no worked example & no & \rd{16}{16} & \rd{15}{16} & \rd{16}{16} & \\
plain copy, no worked example & yes & \rd{16}{16} & \rd{16}{16} & \rd{16}{16} & \\
copy, one worked example in the request & no & \rd{0}{16} & \rd{0}{16} & \rd{0}{16} & \rd{0}{16} \\
copy, one worked example in the request & yes & \rd{12}{16} & \rd{13}{16} & \rd{16}{16} & \rd{16}{16} \\
chained hop, second name given & no / yes & \multicolumn{2}{c}{\rd{4}{7} / \rd{7}{7}} & \multicolumn{2}{c}{\rd{5}{5} / \rd{5}{5}} \\
\midrule
\multicolumn{6}{@{}p{14.6cm}@{}}{description harness, the unit arriving at the model's own naming of it: unmasked in all 32 layers, it draws 0.05--0.08 of the attention a resident unit receives; the verbatim read goes from \rd{1}{81} to \rd{81}{81} once the question is re-encoded over it} \\
\bottomrule
\end{tabular}

\end{table}

\subsection{Values}
\label{sec:values}

Keys decide what is fetched; values decide what is read. Table~\ref{tab:values} collects what a far value must be.
Values are as word-clustered as keys: two values of the same word far apart are more alike than two context neighbors at every depth, by a wide margin in the early and late layers, and a ridge fit on word identity beats a fit on the neighbor mean at every one of the 32
layers. The asymmetry between keys and values is in the consequence of getting them wrong. Far keys reduced to the
token's identity alone cost almost nothing on loss, and read no needles; far values reduced the same way cost a
large fraction of a nat. Replacing every far value in an 8K document with the token's own bare identity code costs
about half of what zeroing them costs; a map fitted on anchor types costs more; permuting the identities costs
several nats, worse than zeroing. The readers consume the per-token code: far values reduced to the token's own
earlier value keep most of the read's worth, the context remainder alone keeps a fifth, its share rising to about a
third in the last layers, and a wrong code costs seven times deletion; the copy-head reader at layer 20 is the most
remainder-driven reader of all, so the mid-stack readers of Section~\ref{sec:what} lean on the remainder more than
the population does. In other words, a wrong far value is worse than no value at all.

\begin{table}[htbp]
\centering\footnotesize
\caption{What a far value must be. \Mistral{} unless stated; the replacement rows are one 8K document with the keys
left real, the read rows $n = 16$ documents.}
\label{tab:values}
\begin{tabular}{@{}p{4.0cm}p{4.6cm}p{7.0cm}@{}}
\toprule
measurement & setting & result \\
\midrule
same-word cosine of values, against context-neighbor cosine & layers 2 / 16 / 31; pairs over 64 tokens apart; chance $\leq 0.01$ & 0.77 / 0.31 / 0.28 against 0.01 / 0.26 / 0.11 \\
loss with far values replaced, keys real & one 8K document; real loss 1.919 nats & zeroed $+0.82$; token-identity code $+0.44$; map fitted on anchor types $+0.71$; identities permuted $+2.8$ to $+3.1$ \\
what the readers consume & far values reduced to the token's own earlier value & 84\% of the read's worth kept ($+0.044$ against $+0.266$ for deletion); context remainder alone 18\%; wrong code $+2.0$ nats, seven times deletion \\
synthesized keys, one vector per token & from one shared layer-8 vector through a per-layer map: cosine 0.85 per word type, 0.98 per occurrence, a shortlist within one to six points of the real keys; per-token keys at cosine 0.94 & needle read \rd{0}{16} on both models \\
synthesized values & keys real; addressing intact (0.7--0.99 of the far mass on the unit) & read \rd{0}{16}; \Llama{} loss: synthesized keys $+0.386$, synthesized values $+0.017$ \\
values rebuilt by a forward over the unit's own tokens & no preceding context; key cosine 0.84 & \rd{16}{16} at 32K, \rd{16}{16} at 128K, \rd{14}{14} at 1M; bulk loss $+0.018$ nats over all 776 eligible units of a 32K document, $+0.0003$ with 256 tokens of preceding context \\
\bottomrule
\end{tabular}

\end{table}

No key or value synthesized outside the model reads.
Every head's key for a word is predictable from one shared layer-8 vector through a
per-layer map at high cosine per word type and higher per occurrence, which ranks a shortlist within a few points of
the real keys and is dead for serving attention: per-token synthesized keys read no needle on either model, and
synthesized values leave the addressing intact and destroy the readout, at a small loss cost against a large one for
synthesized keys. A forward pass over the unit's own tokens with no preceding context, at a lower key cosine than the
map reaches, reads every needle from 32K to 1M: a self-consistent forward pass outperforms a higher-fidelity map. Its bulk cost
over every eligible unit of a 32K document is a few hundredths of a nat with no preceding context and almost nothing
with a few hundred tokens of it. Only the model's own values read, whether written in place or rebuilt by the model
from the text; the asymmetry agrees with the cache-side finding that values tolerate less approximation than
keys~\citep{cui2025asymkv} and with the accuracy cost reported for predicted keys and
values~\citep{horton2024kvprediction}. Multi-token words are assembled into single vectors (Table~\ref{tab:assembly}). A multi-token word
is assembled in the residual at its last piece between layers 2 and 6 on both models: the embedding table's rows
predict human relatedness for pairs of multi-token words far worse than for single-token pairs, and the layer-5 to
layer-6 residual lifts the multi-token pairs to the single-token level while the single-token pairs barely gain. The
assembled vector is order-coded, it is built by the first feed-forward layers, and read through the embedding table
it names a single-token partner far better than the logit lens does.  We speculate that this inability to synthesize keys and values arises from the frame the model chooses during training to operate in~\citep{oskin2026offaxisonpurpose}.  
Synthetic values have to be generated in the right frame, otherwise they appear as noise to the model. Future work may find a way to generate keys and values outside the model by working in that frame.

\begin{table}[htbp]
\centering\footnotesize
\caption{A multi-token word is assembled as a token. Both models; relatedness is Spearman correlation with human
ratings over 6,607 rated word pairs.}
\label{tab:assembly}
\begin{tabular}{@{}>{\raggedright\arraybackslash}p{6.4cm}>{\raggedright\arraybackslash}p{9.6cm}@{}}
\toprule
measurement & result (\Mistral{} / \Llama{}) \\
\midrule
human relatedness predicted from the embedding table's rows, over 6,607 rated word pairs & Spearman 0.15 / 0.06 for pairs of multi-token words, against 0.63 / 0.62 for single-token pairs \\
the same from the layer-5 to layer-6 residual at the word's last piece & multi-token pairs 0.59 / 0.51; single-token pairs gain at most 0.06 over their table rows \\
the assembled vector is order-coded & cosine to the intact vector 0.21 with the pieces reversed and 0.27 shuffled, against a 0.96 single-token ceiling \\
built by the feed-forward layers 1--3 & ablating them takes the best multi-token head's discrimination from 0.92 to 0.62 \\
read through the embedding table & names a single-token partner at a median rank of 7 of 13,571, where the logit lens ranks it 96 \\
\bottomrule
\end{tabular}

\end{table}

\subsection{Far reads must be precise}
\label{sec:descriptions}

\begin{figure}[htbp]
\centering
\includegraphics[width=0.92\textwidth]{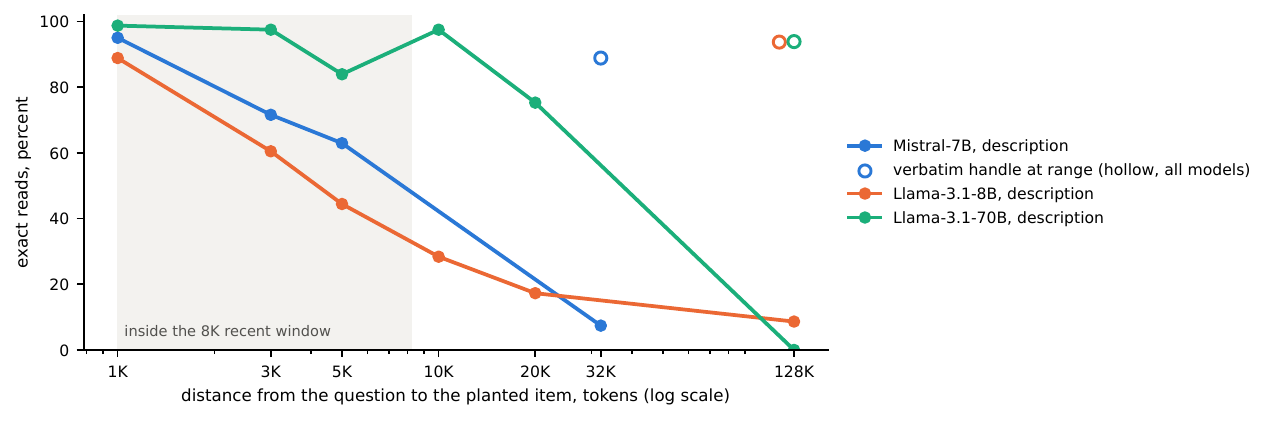}
\caption{Description-matched reads against distance under full attention, three same-format decoys, worked example
in the document tail, $n = 81$ per point ($n = 82$ for the 70B at 128K). \Mistral{} on a Project Gutenberg haystack;
the two Llama models on a control haystack to 20K and, for the far points, the Gutenberg haystack (8B) and the control
haystack (70B). Hollow markers are the verbatim-handle question on the same items at range (the 8B marker is offset
left for visibility). Every model resolves the description at the front of the window and none at range; the word
crosses the window in every model.}
\label{fig:descriptions}
\end{figure}

Figure~\ref{fig:descriptions} and Table~\ref{tab:descriptions} plant a seven-digit code with a handle word and ask for it by description, with no
content word shared between question and item, three same-format decoys present, a one-shot worked example in the
document tail, and full attention over the whole document, all in the PyTorch harness. At the front of the window
the description is resolved: nine times in ten or more a thousand tokens back on every model, falling with distance on the 7--8B models and staying above three quarters to 20K on the 70B. At range the read is gone in every model, the 70B included, with and
without the worked example. The verbatim handle survives the same trip on every model. Scale closes the in-window
gap and the dependence on a worked example: on the control haystack at 3K the 70B reads descriptions far better than
the 8B, with and without a worked example. It does not close the far read. The collapse is a property of the read
rather than of model size, and it is the in-cache form of the collapse NoLiMa measures when the needle shares no
word with the question~\citep{modarressi2025nolima}.

\begin{table}[htbp]
\centering\footnotesize
\caption{Reads at range by description and by word. Description-matched questions with three same-format decoys,
full attention, PyTorch harness, $n = 81$ per point unless stated ($n = 82$ for the 70B at range). \Mistral{} on a
Project Gutenberg haystack; the two Llama models on a control haystack inside the window and, at range, the
Gutenberg haystack (8B) and the control haystack (70B).}
\label{tab:descriptions}
\setlength{\tabcolsep}{4pt}
\begin{tabular}{@{}>{\raggedright\arraybackslash}p{5.4cm}>{\raggedright\arraybackslash}p{3.5cm}>{\raggedright\arraybackslash}p{3.3cm}>{\raggedright\arraybackslash}p{3.0cm}@{}}
\toprule
 & \Mistral{} 32K & \Llama{} 128K & \LlamaBig{} 128K \\
\midrule
\multicolumn{4}{@{}l}{\itshape the description, inside the window (percent read; worked example in the document tail)} \\
1K / 3K / 5K back & 95.1 / 71.6 / 63.0 & 88.9 / 60.5 / 44.4 & 98.8 / 97.5 / 84.0 \\
10K / 20K back & & 28.4 / 17.3 & 97.5 / 75.3 \\
3K back with no worked example (control haystack) & & 8.6 & 81.5 \\
\midrule
\multicolumn{4}{@{}l}{\itshape at range} \\
the description, with / without the worked example & \rd{6}{81} / \rd{9}{81} & \rd{7}{81} / \rd{13}{81} & \rd{0}{82} \\
the verbatim handle & \rd{72}{81} & \rd{15}{16} & \rd{77}{82} \\
\midrule
\multicolumn{4}{@{}l}{\itshape placement and content matching do not rescue the description read} \\
the needle's unit placed in range, three competitors ($n = 25$): the verbatim handle / the description & \rd{24}{25} / \rd{1}{25}, against \rd{20}{25} / \rd{1}{25} under full attention & & \\
far units ranked by a per-head content vector: the described unit placed in range / read & & \rd{27}{27} / \rd{0}{27} & \\
\midrule
\multicolumn{4}{@{}l}{\itshape the word, once the model has produced it (full attention, harness)} \\
a scratchpad that first names the described thing & \rd{9}{81} to \rd{56}{81} & \rd{13}{81} to \rd{61}{81} & \\
given a correct name & & \rd{59}{59} & \\
\bottomrule
\end{tabular}

\end{table}

Placement does not rescue it (Table~\ref{tab:descriptions}). On \Mistral{} at 32K with three competitors, the
needle's unit placed in range reads the verbatim handle better than full attention over the document does, and reads
the description almost never under both: a nearer position with fewer competitors is a better place for the lexical
read than the far document itself, and it does nothing for the description read. Putting the right unit in front of the
model by a content match does not rescue it either: far units ranked by a per-head content vector put the
description's unit in range every time on \Llama{} at 128K and the model never read it. The description route is a
property of the model rather than of where the key sits, and it lives at the front of the window at every scale.
What does cross the window is the word, once the model has produced it: under full attention in the same harness, a
scratchpad that first names the described thing lifts both models from about one in seven to about seven in ten,
and given a correct name \Llama{} reads every code. Section~\ref{sec:concept} turns the name into a retrieval step.

\subsection{Summary}
\label{sec:implications}

Each of the facts above constrains how a long history can be served. Routing, half of the mass, reads within 64
tokens of the query and is the one place where every plane must be right, so the recent past has to be present at
its true positions. The one operation that forecasts is a few mid-stack heads copying an occurrence chosen by
the words it contains, the more sharply the longer the aligned match, so a far span is worth having back for its
words, and a description of it is no address until the model has turned the description into a word. The far match
rides the slow planes and tolerates any position inside the trained range, degrading beyond about twice it, so a far
span can be brought back at whatever position is convenient, while the sink and every anchor must sit inside the
range. The order of a span was written into its keys and values while it was in the window, and the readers copy
values, so a span must come back as the model wrote it; a synthesized value destroys the read. Under competition the
binding forms at the handle's own query, so a far span must be visible before the question is encoded and must stay
while the answer is copied, and a span that arrives later stays unread until the question is encoded again over it.
Section~\ref{sec:design} is those constraints turned into a design, every setting of which the model card fixes.

\secbarrier

\section{Design}
\label{sec:design}

Section~\ref{sec:geometry} describes how attention operates at the geometric level. In this section we use that
information to retrofit unbounded context onto an already trained transformer. The overall design is described in
Figure~\ref{fig:design}. A request's life has three phases. Prefill runs window-only and streams the keys and values
that roll off the near window to the host while the index is built. The hand-off gathers for the request's words and
the seed, places the units at slots, and then encodes the request over them. Decode fetches on every generated rare word, rewinds the cache to the start of the request, and encodes the request and the tokens generated so far again over what arrived.

Each of the design's parts has a purpose (Section~\ref{sec:what}). Routing is served by a near window of dense
attention at true positions. The far read is lexical, so the index is a word index: a unit is fetched for the rare
words it contains. The far read is position-free, so a fetched unit is placed at a slot inside the trained range, at
whatever distance is convenient, and its slow planes carry the match there; that is the fact the whole retrieval tier
is built on: a far unit is worth fetching for the words it contains, and once fetched, its precise position is not
relevant. The reader heads copy an occurrence and what followed it nearby, so a unit's tokens must arrive in the
order the model wrote them, along with that context. The copy is a read of a value, so values are stored, never
synthesized (Section~\ref{sec:values}). Attention relies on a sink token to turn a head off and on the first
tokens of the document as anchors for its softmax floor, so we keep the sink and the first 128 document tokens inside
the trained range at all times (Section~\ref{sec:placement}). Nothing in the design is per head or per layer; the
same far window serves every layer alike.

\subsection{Parameters}
\label{sec:knobs}

Table~\ref{tab:knobs} lists every knob of the design. None needs the user's intervention: the defaults are derived
from the model card of the transformer. Models must be trained with per-plane rotation, so we do not support
multi-head latent attention~\citep{deepseekv2} or a linear-bias scheme~\citep{press2022alibi}, and the design's rule refuses a configuration whose deepest slot would lie outside the trained range, a check the engine does not perform (Section~\ref{sec:implementation}). The one exception to derivation is the plane
count, which we set by measuring \Mistral{} and finding how large a safety margin is needed; the engine does not apply it. The measurements behind the rows are those of Section~\ref{sec:geometry}: the plane count in
Section~\ref{sec:planes}, the range rule and the sink rule in Section~\ref{sec:placement}, the event rule in
Section~\ref{sec:residency}, and the bit widths in Section~\ref{sec:values}.

\begin{table}[htbp]
\centering\scriptsize
\caption{Every knob of the design, derived from the model card, with the measurement behind it. $\W$ is the near
window, $\band$ the slot band, $L$ the trained range. Values are the configuration on \Mistral{} and \Llama{}; we measured the two compression rows in the harness, and the engine stores the host store uncompressed. In the engine $\W$ is a lower bound: prefill above 24,706 tokens runs in 1,024-token batches,
so a query sees $\W$ to $\W + 1{,}023$ recent tokens.}
\label{tab:knobs}
\begin{tabular}{@{}p{2.4cm}p{2.7cm}p{4.5cm}p{5.5cm}@{}}
\toprule
knob & value (\Mistral{} / \Llama{}) & rule, from the model card & the measurement behind it \\
\midrule
trained range $L$ & 32,768 / 131,072 & the hard ceiling: refuse if $\W + \band + \text{prefix} + 2 > L$ & a key presented outside the range is unreadable: the sink at its true distance costs $+0.58$ / $+1.15$ nats at 1M, at a slot the loss is flat to 1M (Section~\ref{sec:placement}) \\
near window $\W$, slot band $\band$ & 8,192 / 16,384 & split the context budget $\W : \band = 1 : 2$; with one given derive the other and re-check against $L$ & the measured setting; halving the band at 128K costs single needles (0.875 against 1.000, $n = 16$) \\
pinned prefix & 128 & the first 128 tokens of the document, at slots & the measured setting; at true distance the prefix costs 3.9--5 nats at 1M \\
key planes kept & 28 of 64 (harness, \Mistral{}; the engine stores keys uncut; \Llama{} rungs unmeasured) & the analytic cut keeps every plane coherent across $\W$ ($\lambda \geq 2\pi\W$ at the effective per-plane rate, 22 planes on \Mistral{}); the default adds one measured rung of margin, a calibration rather than a derivation & three-key discrimination against planes kept (Section~\ref{sec:planes}, harness, $n = 16$): \rd{16}{16} down to 22 planes, \rd{14}{16} at 15; the analytic cut $\lambda \geq 2\pi \W$ keeps 22, the default 28 keeps one rung of margin \\
key and value bits & 4 / 4, groups of 32 (measured; the engine stores f16) & the smallest width that is free & bits are free: four-bit keys and values read the same single and three-key rows as eight-bit and bf16, loss within 0.0025 nats across the sweep \\
key centering & per head & the per-head mean over the first $\W$ tokens, then frozen & measured free in the harness; $\W$ is the span the window holds anyway \\
sinks & BOS; on \Mistral{} the first newline & BOS always; any other token whose layer-2 residual norm is over 50 times the running median & BOS is 614 times and the first newline 712 times the median token at layer 2; the second, third and tenth newlines and the median of 502 newlines are about 1 times \\
admission & 15 nats & a word enters the index only above 15 nats of surprisal on the running unigram model; a valve raises the threshold when demand exceeds $\band/2$ & demand at 1M falls from 200,002 to 1,700--2,800 tokens per gather, coverage from 0.08 to about 1.0; the needle unit ranks first among the hits; 32K unchanged \\
occurrences kept & 16 per word & keep the last 16, oldest dropped & bounds demand by the query, words $\times$ 16 $\times$ unit length, about 8,000 tokens at any length \\
units & sentences & whole words from the tokenizer's piece boundaries; sentence units for every source & sentence units win or tie every needle cell at far less band traffic (about 700 tokens per request against 9--14K for window units with the far seed off; 1,062--1,768 at 32K and 128K, 2,342--3,058 at 1M, in the rows of Section~\ref{sec:results}) \\
events & 15 nats, budget 2,048 & the same threshold on generated whole words; one exact re-encode of the request and generated tail per event, within a per-turn token budget & a unit arriving mid-generation is read only after the re-encode (Section~\ref{sec:residency}); the gate loses no read \\
sliding-window layers & window only & a layer that cannot see past its own window gets no band & by construction \\
out of class & refuse & MLA, and non-rotary schemes such as ALiBi & slot placement re-rotates keys plane by plane; neither has the per-plane rotation that makes a slot meaningful~\citep{deepseekv2,press2022alibi} \\
\bottomrule
\end{tabular}

\end{table}

\subsection{Dynamically constructed window}
\label{sec:design:range}

The near window ($\W = 8{,}192$ tokens) of the attention window is stored densely and in the order the transformer
would normally see it. The sink and the pinned prefix of the first 128 document tokens are kept for the anchoring
mass of Section~\ref{sec:what}, and kept at slots inside the trained range. What goes into the remainder of the
window, the far window of $\band = 16{,}384$ slots, we vary dynamically. Every cell the model sees sits at a position
inside the trained range, because a key presented beyond the range is unreadable (Section~\ref{sec:placement}): a
fetched key is presented just beyond the near window regardless of where it really was, and inside the range that
placement is free. Table~\ref{tab:design} holds the measurements behind this section.

\begin{table}[htbp]
\centering\scriptsize
\caption{The measurements behind the design. Reads are exact matches of $n$; harness rows are the PyTorch harness on
\Mistral{} at $n = 16$, and the rest are the \llamacpp{} engine. Section~\ref{sec:geometry} holds the measurements
of attention the rows rest on, and Section~\ref{sec:results} the results.}
\label{tab:design}
\setlength{\tabcolsep}{4pt}
\begin{tabular}{@{}>{\raggedright\arraybackslash}p{2.4cm}>{\raggedright\arraybackslash}p{4.9cm}>{\raggedright\arraybackslash}p{8.1cm}@{}}
\toprule
part & what we measured & result \\
\midrule
\multicolumn{3}{@{}l}{\itshape the host store and its index} \\
store size & keys and values of every far token at f16, 32 layers & 131 KB per token; 128.5 GiB at one million tokens on \Mistral{} \\
admission at 15 nats, last 16 occurrences & tokens a gather asks for at 1M, and coverage of the planted item & 200,002 tokens per gather without the two rules, 1,700--2,800 with them; coverage 0.08 to about 1.0, the item's unit ranking first among the hits; nothing changes at 32K \\
sentence units against $\pm 256$ / $\pm 512$ windows & needle cells, band traffic, bulk loss, RULER volume tasks & sentence units win or tie every needle cell (two-hop on \Llama{} at 128K 84 against 66--68) at about a tenth of the band traffic; windows win bulk loss by 0.009 nats on code and 0.001--0.004 on prose, and the volume subset of RULER on \Mistral{}, 55.7 against 47.6 \\
whole words against fixed chunk boundaries & single needles, harness & \rd{14}{16} when the chunk cuts the handle, \rd{16}{16} with whole words \\
compression of the key planes (harness, \Mistral{}) & three-key discrimination against planes kept; bytes per token & \rd{16}{16} down to 22 planes, \rd{14}{16} at 15; the analytic cut (wavelength at least $2\pi\W$) keeps 22, the default 28 keeps one rung of margin; 29.0 KB per token at 28 planes and four bits, against 33.0 at 40, 27.5 at 22, 24.8 at 15, and 131 KB uncompressed; loss moves by less than 0.0005 nats across the ladder \\
a store budget & usage-at-exit eviction of whole units on a 1M document, capped at 512K and 256K tokens & single needles \rd{8}{16} and \rd{5}{16}, three-key \rd{2}{16} and \rd{2}{16}, against \rd{16}{16} unbounded; every evicted unit had never been used \\
\midrule
\multicolumn{3}{@{}l}{\itshape the far window} \\
slot placement & fetched keys at a slot just beyond the near window, against their true distance & free inside the trained range (Section~\ref{sec:placement}); at its true distance the sink costs $+0.58$ / $+1.15$ nats at 1M and the prefix 3.9--5 nats, at slots the loss is flat to 1M; fetched keys scored at their true distance beyond 128K read worse than the near window alone at every length \\
fill per request & tokens fetched, of 16,384 slots & 1,062--1,768 at 32K and 128K, 2,342--3,058 at 1M; the back-fill covers about two thirds of the far region at 32K, 16K of 123K far tokens at 128K, one in sixty at 1M \\
hand-off before the request is encoded & the identical gather, visible from the start of the request or only after the handle word & \rd{53}{56} against \rd{0}{56} (\rd{78}{81} at $n = 81$; \rd{52}{55} with a 300-token worked example in the request); in \llamacpp{} at 128K \rd{16}{16} against \rd{13}{16} single needles \\
keep-fill: topping up from the far window's own contents & single needles; the ring's advance per gather & at 32K 0.812 to 1.000 on both models, at 128K 0.875 and 0.938 to 1.000; the advance 2,396 to 1,046 cells \\
the far seed & two-hop; the handle being seeded; three-key & $+3$--13 points; the handle seeded in 97--100 of 100 requests, from 76--82; three-key within 3 \\
the far window through generation & a unit evicted before the digits are produced & \rd{0}{16}, in the harness and in \llamacpp{} alike \\
the always-full far window & the description read through naming on \Mistral{} (Section~\ref{sec:concept}) & \rd{36}{81} on the engine against \rd{51}{81} with only the fetched units resident; denying the band the same-format decoys recovers 9 of the 15 points \\
\midrule
\multicolumn{3}{@{}l}{\itshape events and prefill} \\
an event's re-encode & tokens covered, time on one H200, exactness & 30--80 tokens, 31--37 ms; byte-identical in its logits to a fresh prefill; the unit first affects a prediction two tokens after the word closes \\
the surprisal gate at 15 nats, the per-turn budget & reads in the worked-example cells ($n = 16$); re-encodes per 128 decoded tokens & \rd{12}{16}, \rd{13}{16}, \rd{16}{16}, \rd{16}{16}, no read lost; 0 / 0 / 4 on \Mistral{} and 0 / 1 / 3 on \Llama{} at 8K / 32K / 128K, from three to five without the gate; budget 2,048 tokens per turn \\
window-only prefill & loss against streamed prefill; speed against full-depth prefill & $-0.0001$ nats on prose and $-0.00007$ on 1.1M tokens of code with cross-file references; full-depth prefill is 20~percent slower at 128K and 42~percent at 1M \\
\bottomrule
\end{tabular}

\end{table}

\paragraph{The host store.}
Every token that leaves the near window keeps its real keys and values on the host, 131 KB per token at f16 across
the 32 layers. The store is unbounded, and nothing is evicted from it: attempts to limit it did not work, because it is not possible to know what a later request will ask for. The host side of the design is the frame of the offloaded retrieval systems as well~\citep{xiao2024infllm,liu2024retrievalattention,chen2025retroinfer,lee2025infinitehip}; what differs here is what finds a unit in the store and where the unit is placed.

\paragraph{Compression.}
The geometry of attention points at an obvious compression of the host store. At distance not all planes are
readable; the fast ones appear as noise to the model. Analytically only 22 planes are meaningful for \Mistral{}
(Section~\ref{sec:planes}), and we retain 28 to provide a buffer. Keeping those planes at four bits shrinks the store
per token to about a fifth of its uncompressed size, 131 KB to 29 KB. We measured the cut in the harness on
\Mistral{} and validated it on three-key discrimination rather than on loss, since loss barely moves across the whole
ladder while the three-key read is what moves. The engine stores its keys and values uncut, and \Llama{}'s
rungs are not measured.

\paragraph{The word index.}
The store is indexed by word rather than by token, because the far read is lexical. Words are whole words at the
tokenizer's piece boundaries; a word enters the index only if it is uncommon, above 15 nats of summed surprisal on a
running unigram model of the document, and the index keeps a word's last 16 occurrences. The two rules bound the
demand of a gather by the query rather than by the document: at one million tokens a gather asks for about a
hundredth of the tokens it would ask for without them, and the planted item's unit ranks first among the hits. The
unit around each occurrence is its sentence, so that a handle is never cut at a chunk boundary; sentence units read
every needle cell at about a tenth of the band traffic of fixed windows, which win only on bulk loss and on the
volume tasks of RULER, and a window of $\pm 256$ or $\pm 512$ tokens around each occurrence is a user flag for code
and volume-heavy use.

\paragraph{What to put in the far window.}
\label{sec:design:store}
At the hand-off we scan the request for words that are in the index, and add the seed: the 32 rarest words of the
near window's last 4,096 tokens that have an occurrence beyond it, at half weight. The seed fetches the far units of
rare words the near window already mentions, which is the first hop of a two-hop question, and it lifts two-hop reads
without costing three-key discrimination. A unit is scored by the summed surprisal of the distinct query words it
matches, and units are taken in score order, ties by recency. The gather happens before the request is encoded,
because a far unit is read only if it is visible at the query of the question's own handle word
(Section~\ref{sec:residency}): the identical gather reads nearly every needle when its units are visible from the
start of the request and none when they arrive after the handle word. The far window is always full. It is filled in
three layers: the units fetched for this request; then whatever it already held, newest first; then a back-fill of
the most recent far tokens down from the near window's edge. Slots are ranked by recency, newer nearer the near
window. At every length the fetched units are a small part of the far window, so most of it is back-fill.

\paragraph{Events.}
\label{sec:design:events}
When the model generates a rare word, its postings are fetched and written into the far window; the unit first
affects a prediction two tokens after the word closes. Retrieval that fires on what the model is generating is the
active-retrieval idea~\citep{jiang2023flare,su2024dragin}, here inside the model's own cache. A unit that arrives mid-generation is not read until the request and the generated tail are encoded again with it present (Section~\ref{sec:residency}), so after each fetch the engine rewinds the cache to the start of the request and encodes the request and the tokens generated so far again, with the fetched units in the far window, and then continues generating. This re-encode is exact, byte-identical in its logits to a fresh prefill, and a few tens of milliseconds long. The event is gated on the generated word's surprisal at the same 15
nats as admission, which loses no read and holds the re-encodes to a few per hundred generated tokens, and it carries
a per-turn budget, since tokens can regain attention late in a long reasoning trace~\citep{zhang2025lazyeviction}.

\paragraph{Prefill.}
Prefill runs window-only. Second hops are generation-time work, and no binding forms at prefill time across units
under any prefill regime (Section~\ref{sec:residency}); gathers during prefill add nothing measurable to loss, and
full-depth prefill is slower by a fifth at 128K and by two fifths at 1M. Above 24,706 tokens prefill runs in
1,024-token batches with the near window rolled per batch, so a query sees between $\W$ and $\W + 1{,}023$ recent
tokens, and $\W$ is a lower bound.

\paragraph{Eviction.}
The far window evicts at every gather, in the reverse of
the order it is filled: the back-fill of recent far tokens goes first, then the oldest of what the window held, and
the units fetched for a request stay through its generation, because a unit evicted before the digits are produced
is never read. Topping up from the window's own contents before back-filling keeps a unit fetched once from being the
first thing overwritten, and it is what takes single needles to a perfect read at 32K and 128K on both models.

\secbarrier

\section{Implementation}
\label{sec:implementation}

The engine's share of the work is small, because the model does the reading; the engine moves cells and maintains the
store and the index. We implemented the design in \llamacpp{}~\citep{llamacpp} as one small change to the library and a driver that
holds every policy. The library gains an entry point that moves a cache cell to a new position, re-rotating its key
by the difference, together with read and write accessors for a cell's key and value and an accessor for the rotary
base; nothing else in the library changes, and with the design off the binary is the stock engine. The driver
captures the keys and values that roll off the window and streams them to the host, builds the word index as the
document is prefilled, gathers and places units at the hand-off and at events, and re-encodes the request and the
generated tail over a late arrival. Every knob of Table~\ref{tab:knobs} is derived from the model card at startup
and printed. The models are the public Q8\_0 files of \Mistral{} and \Llama{}, the latter converted from the
original weights with its rotary frequency table; f16 weights read the same strings as Q8\_0 on 15 of 16 examples.

We test the engine against itself, and Table~\ref{tab:tests} holds the checks and the costs. Below the trained range
the engine is byte-identical to the stock engine. Every cell of the far window and the near window carries its
provenance, and the provenance checks find no bad cell. The pinned prefix, re-placed from the host at every move,
stays at its true keys through the moves of a 1M prefill, where a prefix moved in place drifts. The re-encode at an
event is byte-identical in its logits to a fresh prefill on both models. The performance path, which feeds prefill
in batches of 1,024 tokens, captures the previous block to the host while the next computes, and captures decode per
block, is byte-identical to the one-block-at-a-time path on nearly every regression text and changes no exact-match
count. Two tests are not implemented: refusing a configuration whose deepest slot lies outside the trained range,
and resuming a session's store. The costs are accounted per token and per request in the same table. Two parts of the
implementation were measured separately: the placement path and the sink over a million tokens.

\begin{table}[htbp]
\centering\scriptsize
\caption{The engine checked against itself, and its costs. The sink rows are the diagnosis of Section~\ref{sec:impl:onemillion}.}
\label{tab:tests}
\setlength{\tabcolsep}{4pt}
\begin{tabular}{@{}>{\raggedright\arraybackslash}p{4.6cm}>{\raggedright\arraybackslash}p{4.6cm}>{\raggedright\arraybackslash}p{6.4cm}@{}}
\toprule
check & setting & result \\
\midrule
\multicolumn{3}{@{}l}{\itshape against the stock engine} \\
logits below the trained range & 9,320, 16,448 and 24,320 tokens & 0 of 32,768 logits differ \\
provenance of every cell & band and window cells after moves & 0 bad of 786,432 band cells and 393,216 window cells \\
the pinned prefix after the moves of a 1M prefill & re-placed from the host at every move / moved in place & cosine 0.999999 / 0.992304 to its true keys \\
the re-encode at an event & both models & byte-identical in its logits to a fresh prefill \\
the batched performance path & 64 regression texts, against one block at a time & byte-identical on 62 of 64; no exact-match count changes \\
\midrule
\multicolumn{3}{@{}l}{\itshape placement by re-rotation} \\
re-rotated capture against pre-rotary capture & a fixed binary, 16 examples at 1M and at 128K & identical strings on all 16 \\
single needle and three-key under either capture path & both models at 128K, $n = 16$ per cell & \rd{16}{16} in all eight cells, identical per-example digit shares \\
the pre-rotary capture path & prefill cost & 25~percent of prefill for no accuracy; off \\
\midrule
\multicolumn{3}{@{}l}{\itshape the sink over a million tokens (\Mistral{}, window-only prefill)} \\
moves of the sink and pinned prefix & one f32 rotation rounded to f16 per block & about 8,100 moves by one million tokens \\
a 128-position turn in the slowest planes & against half an f16 unit in the last place & $1.6 \times 10^{-4}$ radians against $4.9 \times 10^{-4}$: those planes stop turning \\
$1 - \cos$ between the sink key and its true value & at 128K / at 1M & mean 0.0023 / 0.19 ($\times 80$ for $\times 8$ shifts); cosine 0.432 in all 16 examples at 1M \\
the fix's cost & prefill at 128K on one H200; decode & 4,574 against 4,772 tokens per second (4~percent); decode unchanged \\
\midrule
\multicolumn{3}{@{}l}{\itshape costs per token and per request} \\
host memory & f16 & 131 KB per token \\
GPU cache per layer & window, far window, pinned prefix & $\W$ to $\W + 1{,}023$ window cells, 16,384 far-window cells, 129 pinned cells \\
the hand-off & 32K and 128K & about 0.2 s \\
re-encodes & per 128 generated tokens & 0 to 4 (Table~\ref{tab:perf}) \\
\bottomrule
\end{tabular}

\end{table}

\subsection{Placement by re-rotation}
\label{sec:impl:rerotation}

\llamacpp{} stores keys already rotated at their position and re-rotates them by a shift when they move; a pre-rotary
store keeps keys unrotated and rotates them once at placement. The two are equivalent, as Section~\ref{sec:placement}
measures and Table~\ref{tab:tests} checks on the engine: on a fixed binary the two paths produce identical strings at
1M and at 128K; on both models at 128K, single needle and three-key discrimination read in full under either capture
path, with identical per-example digit shares; and a shift of the whole sequence by up to a million positions leaves
the decoded tokens and the dense logits where they were (Table~\ref{tab:placement}). The engine re-rotates: a pre-rotary store would cost a quarter of prefill for the same reads, and compression does not need it, since a plane cut commutes with rotation.

\subsection{Keeping the sink exact over a million tokens}
\label{sec:impl:onemillion}

The engine re-places the pinned prefix from the host at every move, one exact rotation each, because a prefix moved in place accumulates rounding error. Window-only prefill moves the sink and the pinned prefix up one block at a time, and each move is an f32 rotation rounded to f16, thousands of times by one million tokens. In the slowest planes a 128-position turn is under half an f16 unit in the last place, so an in-place prefix stops turning in those planes; the angle error grows linearly in the number of moves and the cosine loss quadratically, and by one million tokens an in-place sink key has drifted far from its true value (Table~\ref{tab:tests}). Every token encoded late in such a prefill is computed against a drifted floor, and on \Mistral{} no single needle is read at one million tokens with the needle resident. Re-placement keeps the sink at its true value and reads every needle (Table~\ref{tab:onemillion}), at a cost of a few percent of prefill at 128K and nothing on decode. \Llama{} reads every needle at 1M without it.

\begin{table}[H]
\centering\footnotesize
\caption{The sink over a million tokens, the same 16 documents throughout, one seven-digit needle each. We measure the sink
key's cosine to its true value before the request. The rows are the prefix moved in place, rewritten once before the question, rewritten by a hand-off before the question, and re-placed from the host at every move.}
\label{tab:onemillion}
\begin{tabular}{@{}p{2.2cm}p{2.7cm}p{4.9cm}p{3.8cm}r@{}}
\toprule
model & hand-off & pinned prefix during prefill & sink key cosine to its true value & reads \\
\midrule
\Mistral{} & after the question & moved in place & 0.432 (all 16 examples) & \rd{0}{16} \\
\Mistral{} & after the question & moved in place, rewritten once before the question & 0.99997 at the question & \rd{7}{16} \\
\Mistral{} & before the question & moved in place & 0.432, then rewritten by the hand-off & \rd{9}{16} \\
\Mistral{} & after the question & re-placed from the host at every move & 0.99997 & \rd{12}{16} \\
\Mistral{} & before the question & re-placed from the host at every move & 0.99997 & \rd{16}{16} \\
\Llama{} & before the question & moved in place & not measured & \rd{16}{16} \\
\bottomrule
\end{tabular}

\end{table}

\secbarrier

\section{Results}
\label{sec:results}

We measure the implementation in \llamacpp{} on stock model files, on one H200, with the question placed at the end of a long document.  Three planted tasks and one public benchmark carry the
numbers. A \emph{single needle} is a seven-digit code introduced by a handle word (``the secret code for the \ldots{}
is''), planted far back and asked for by its handle, the arrival test of the needle protocol~\citep{kamradt2023niah}.
\emph{Three-key discrimination} plants three same-format codes and asks for one by its handle: the discrimination
test, the one that competition breaks. A \emph{two-hop} question names an item whose far unit names the code, with the
bridge sentence inside the window: the answer needs a first hop (produce the bridge word) and then the far read; we
score the bridge word as well as the code. Every planted row has $n = 100$ documents from one item seed and we read
it with a Wilson 95~percent interval; within one seed, a difference inside the intervals is parity, never a gain. Two
references bound each row: dense attention (stock \llamacpp{} with the whole cache resident, where it fits) and the
window-only floor (the near window and the pinned prefix with no retrieval). We run RULER~\citep{hsieh2024ruler} from
its official generator, 13 tasks at $n = 100$ each, scored by its own string-match metric, and measure throughput
against stock llama-bench in the same job. A second floor, run where the comparison needs it, holds a 24,576-token window with no retrieval: the design's whole
GPU budget of window and far window, spent on recency.

\subsection{Single needles and three-key discrimination}
\label{sec:results:needles}

Table~\ref{tab:needles} is the arrival and discrimination matrix. Single needles read at or near every one of the
hundred at every length on both models, at parity with dense attention where dense attention fits. Three-key
discrimination reads in the high eighties to the mid nineties, and its differences from dense attention sit inside
or at the edge of the intervals, so they are read as parity. Dense \Mistral{} at 128K reads nothing, because 128K is
four times its trained range, so the design's reads there are reads dense attention cannot make rather than gains
over it. The window-only floor reads a fifth to a third of the needles at 32K, where some fall inside the window, and almost none at 128K; the
matched floor, the design's whole GPU budget spent on recency, reads about one in ten, so the read comes from the far
window rather than from the size of the budget. At one million tokens nothing is lost from 128K, with a few thousand tokens fetched per request. The PyTorch harness agrees in every one of its cells from 32K to 1M on both models,
where a sliding window with its sink kept reads no single needle at any length.

\begin{table}[htbp]
\centering\footnotesize
\caption{Single needles and three-key discrimination, exact matches of 100 with Wilson 95~percent intervals, one item
seed. Dense is stock \llamacpp{} with the whole cache; the floor is the near window and pinned prefix with no
retrieval; the matched floor is a 24,576-token window, the design's whole GPU budget, with no retrieval, run at 128K.
$^{\dagger}$Dense \Mistral{} at 128K is four times its trained range. $^{\ddagger}$A dense 1M cache does not fit one H200.}
\label{tab:needles}
\setlength{\tabcolsep}{3.5pt}
\begin{tabular}{@{}llrrrrrr@{}}
\toprule
task & model & context & design & fetched tokens & dense & floor & matched floor \\
\midrule
single needle & \Mistral{} & 32K & 99 [95--100] & 1,142 & 99 [95--100] & 18 [12--27] & \\
single needle & \Mistral{} & 128K & 100 [96--100] & 1,768 & 0 [0--4]$^{\dagger}$ & 0 [0--4] & 11 [6--19] \\
single needle & \Llama{} & 32K & 99 [95--100] & 1,062 & 100 [96--100] & 18 [12--27] & \\
single needle & \Llama{} & 128K & 100 [96--100] & 1,604 & 100 [96--100] & 0 [0--4] & 11 [6--19] \\
three-key & \Mistral{} & 32K & 88 [80--93] & 1,132 & 93 [86--97] & 31 [23--41] & \\
three-key & \Mistral{} & 128K & 93 [86--97] & 1,759 & 0 [0--4]$^{\dagger}$ & 1 [0--5] & 9 [5--16] \\
three-key & \Llama{} & 32K & 96 [90--98] & 1,090 & 98 [93--99] & 31 [23--41] & \\
three-key & \Llama{} & 128K & 94 [88--97] & 1,632 & 87 [79--92] & 1 [0--5] & 8 [4--15] \\
\midrule
single needle & \Mistral{} & 1M$^{\ddagger}$ & 100 [96--100] & 3,058 & does not fit & & \\
single needle & \Llama{} & 1M$^{\ddagger}$ & 100 [96--100] & 2,342 & does not fit & & \\
three-key & \Mistral{} & 1M$^{\ddagger}$ & 99 [95--100] & 2,757 & does not fit & & \\
three-key & \Llama{} & 1M$^{\ddagger}$ & 100 [96--100] & 2,353 & does not fit & & \\
\bottomrule
\end{tabular}

\end{table}

\subsection{Two hops}
\label{sec:results:twohop}

Table~\ref{tab:twohop} is the two-hop matrix. \Llama{} reaches parity with dense attention at 128K and sits seven points below it at 32K. \Mistral{} sits well below dense attention at 32K and reads a little over half at 128K, where
dense attention is out of range. The bridge-word column locates the loss. On \Mistral{} at 32K the bridge word is
produced about as often with the needle resident as by the window-only floor, and far more often under dense
attention: the near window alone loses most of the first hop, the far window costs a little more, and the code
follows the word. At 128K the far window's share of the loss grows, and the matched floor produces the word often
and the code never. \Llama{} names the bridge word less often than its own floor does, yet answers nearly as often as
dense attention: it reaches the far code without producing the bridge word, so the first-hop reading bounds
\Mistral{}'s result and not \Llama{}'s. Retrieval is not the bound in either case: the surprisal gate never fired,
since the handles sit far above the admission threshold, and no needle was cut.

\begin{table}[htbp]
\centering\footnotesize
\caption{Two-hop questions of 100: the code produced, and the bridge word produced, for the design, dense attention,
the window-only floor and, at 128K, the matched floor of a 24,576-token window with no retrieval. $^{\dagger}$Dense
\Mistral{} at 128K is out of range.}
\label{tab:twohop}
\begin{tabular}{@{}llrrrrr@{}}
\toprule
model & context & design & dense & floor & matched floor & fetched tokens \\
 & & code / word & code / word & code / word & code / word & per request \\
\midrule
\Mistral{} & 32K & 66 / 73 & 91 / 95 & 0 / 75 & & 1,080 \\
\Mistral{} & 128K & 59 / 59 & 0 / 0$^{\dagger}$ & 0 / 78 & 0 / 84 & 1,759 \\
\Llama{} & 32K & 87 / 46 & 94 / 85 & 0 / 64 & & 1,104 \\
\Llama{} & 128K & 92 / 52 & 92 / 89 & 0 / 72 & 0 / 68 & 1,610 \\
\bottomrule
\end{tabular}

\end{table}

\subsection{RULER}
\label{sec:results:ruler}

Table~\ref{tab:ruler} holds the 13 RULER tasks on \Llama{} at 128K and \Mistral{} at 32K. On \Llama{} the design's
aggregate is a point above dense attention, far above the window-only floor and well above the matched floor, the
design's whole GPU budget spent on recency. The single-needle tasks are at or near perfect, except the third on \Mistral{}, where both arms fall to the seventies. The
multi-key tasks, the stress tasks, read above dense attention on the first two and a few points below it on the
third, where the design misses the usual bar of 50 by three points~\citep{hsieh2024ruler}; multi-value and multi-query
read above dense attention by about ten points. The design loses the tasks that chain through content, variable
tracking and the first question-answering task, the same limit as the first hop above. On \Mistral{} at 32K the
aggregate is about eight points below dense attention, and the largest losses are the second and third multi-key tasks, retrieval under competition, and the two volume tasks, common and frequent words, where the window-unit flag recovers part of the loss (Table~\ref{tab:design}). The multi-key losses are the far-window
competition of Section~\ref{sec:design:store} on a public task: a key with three same-format competitors is the one
read the always-full far window hurts. The remaining controls the RULER bar asks for, a random store and the
published training-free systems, are not in this table.

\begin{table}[htbp]
\centering\footnotesize
\caption{RULER, 13 tasks at $n = 100$ from the official generator, RULER's string-match metric $\times 100$. Dense and
floor as above; $^{\ast}$the matched floor is a 24,576-token window, the design's whole GPU budget of window and far
window, with no retrieval.}
\label{tab:ruler}
\setlength{\tabcolsep}{4pt}
\begin{tabular}{@{}lrrrrr@{\hspace{1.5em}}rrrr@{}}
\toprule
 & \multicolumn{5}{c}{\Llama{} at 128K} & \multicolumn{4}{c}{\Mistral{} at 32K} \\
\cmidrule(lr){2-6}\cmidrule(lr){7-10}
task & design & dense & floor & matched$^{\ast}$ & design $-$ dense & design & dense & floor & design $-$ dense \\
\midrule
single needle 1 & 100.0 & 100.0 & 2.0 & 18.0 & $+0.0$ & 100.0 & 100.0 & 22.0 & $+0.0$ \\
single needle 2 & 98.0 & 100.0 & 8.0 & 19.0 & $-2.0$ & 100.0 & 100.0 & 28.0 & $+0.0$ \\
single needle 3 & 100.0 & 90.0 & 5.0 & 15.0 & $+10.0$ & 77.0 & 72.0 & 21.0 & $+5.0$ \\
multi-key 1 & 99.0 & 87.0 & 8.0 & 18.0 & $+12.0$ & 96.0 & 93.0 & 28.0 & $+3.0$ \\
multi-key 2 & 71.0 & 59.0 & 5.0 & 19.0 & $+12.0$ & 58.0 & 76.0 & 22.0 & $-18.0$ \\
multi-key 3 & 47.0 & 53.0 & 3.0 & 9.0 & $-6.0$ & 53.0 & 73.0 & 17.0 & $-20.0$ \\
multi-value & 91.3 & 78.5 & 8.0 & 16.8 & $+12.8$ & 94.8 & 99.2 & 23.2 & $-4.5$ \\
multi-query & 96.5 & 86.2 & 9.2 & 18.3 & $+10.3$ & 89.0 & 94.5 & 23.2 & $-5.5$ \\
variable tracking & 62.0 & 82.2 & 14.2 & 24.4 & $-20.2$ & 78.2 & 88.8 & 26.2 & $-10.6$ \\
common words & 0.7 & 0.6 & 0.9 & 0.8 & $+0.1$ & 3.8 & 26.6 & 2.5 & $-22.8$ \\
frequent words & 76.0 & 74.7 & 77.7 & 73.7 & $+1.3$ & 59.7 & 79.7 & 44.7 & $-20.0$ \\
question answering 1 & 41.0 & 60.0 & 20.0 & 22.0 & $-19.0$ & 37.0 & 39.0 & 19.0 & $-2.0$ \\
question answering 2 & 38.0 & 37.0 & 31.0 & 31.0 & $+1.0$ & 35.0 & 40.0 & 32.0 & $-5.0$ \\
\midrule
aggregate (13 tasks) & \textbf{70.80} & 69.86 & 14.77 & 21.91 & $+0.94$ & \textbf{67.80} & 75.52 & 23.76 & $-7.72$ \\
\bottomrule
\end{tabular}

\end{table}

\subsection{Throughput and memory}
\label{sec:results:perf}

Table~\ref{tab:perf} sets the design against stock \llamacpp{} on one H200. Below the trained range the design costs under ten percent of prefill and of steady decode, from the copy that streams rolled-off keys and values to the host and
from the window-only prefill above 24.7K tokens; the target for that regime is parity, and it is not met. At 128K
the design is faster than the dense model it runs on, nearly twice the prefill and about two thirds more steady
decode, because every query attends over about 25K cells instead of 128K. Re-encodes at events are rare below 128K
and a few per 128 decoded tokens there. The GPU cache is bounded by construction at the window, the far window and
the pinned prefix, about 25K cells per layer at any length; the host store grows with the document, at 131 KB per
token.

\begin{table}[htbp]
\centering\scriptsize
\caption{Throughput against stock \llamacpp{} on one H200: prefill of the whole context and steady decode after it,
tokens per second, with the design's ratio to stock, and the memory the design holds. Stock \llamacpp{} does not fit a
1M cache on one H200.}
\label{tab:perf}
\begin{tabular}{@{}llrrrrrr@{}}
\toprule
model & context & stock prefill / decode & design prefill & ratio & design decode & ratio & re-encodes / 128 \\
\midrule
\Mistral{} & 8K & 11,017 / 234.1 & 10,324 & 0.94 & 228 & 0.97 & 0 \\
\Mistral{} & 32K & 8,850 / 180.2 & 8,084 & 0.91 & 184 & 1.02 & 0 \\
\Mistral{} & 128K & 4,819 / 91.1 & 8,920 & 1.85 & 150 & 1.65 & 4 \\
\Llama{} & 8K & 11,003 / 226.8 & 10,319 & 0.94 & 214 & 0.95 & 0 \\
\Llama{} & 32K & 8,835 / 175.4 & 8,048 & 0.91 & 161 & 0.92 & 1 \\
\Llama{} & 128K & 4,833 / 89.4 & 8,938 & 1.85 & 150 & 1.68 & 3 \\
\midrule
\Mistral{} & 1M & does not fit & 8,222 & & 149 & & 1 \\
\midrule
\multicolumn{8}{@{}p{14.6cm}@{}}{memory: the GPU cache is 24,705--25,728 cells per layer at every length (window, far window and pinned prefix); the host store is 131 KB per token at f16, 128.5 GiB of keys and values at 1M, and the process peaks at 132.6 GiB on the 1M needle runs} \\
\bottomrule
\end{tabular}

\end{table}

\subsection{One million tokens}
\label{sec:results:onemillion}

At one million tokens, 32 times \Mistral{}'s trained range and eight times \Llama{}'s, the design reads every single
needle and all but one three-key discrimination, fetching a few thousand tokens per request. The engine prefills
the million tokens at 8,222 tokens per second and decodes at 149, with a re-encode every hundred-odd generated tokens,
and the host store holds the million tokens' keys and values in about 130 GiB. The sink must be kept exact through the moves of
a million-token prefill (Section~\ref{sec:impl:onemillion}).

\secbarrier

\section{Matching on descriptions}
\label{sec:concept}

Section~\ref{sec:descriptions} showed that attention cannot find anything but a word in the distant past of the
attention window. A description does match, however, in the near window, and the word that match produces can
reach far back into the history. We test this by asking the model to name the described thing before answering:
the name it produces is a word in the near window, and that word is what reaches far back. Under dense attention
both models produce the name in about three quarters of the cases, and the description read rises from about one in
seven to about seven in ten (Table~\ref{tab:concept}).

Under the design the same question takes four steps, because the sentence that holds the answer is not in the cache
when the question is asked. The model generates the name. The name is a rare word, so the word index fetches the
sentences it occurs in from the host store into the far window, the fetched sentence with the code among them. At
that point the sentence is in the cache but unread: the request and the name were encoded before the sentence
arrived, and Section~\ref{sec:residency} showed that under competition a far unit is read only if it was present
when the question's handle word was encoded. So the engine rewinds the cache to the start of the request and rolls
forward again over the request and the tokens generated so far, with the fetched sentences present. That pass is
exact, byte-identical in its logits to a fresh prefill, and the code the model then produces is read from the
fetched sentence. The harness shows the steps apart. With the far window holding only fetched units, the fetch
without the roll-forward leaves \Mistral{} at almost nothing, since the sentence arrives among three same-format
decoys; the roll-forward reads nearly two thirds; and with the fetch disabled both models read nothing, because the
name is produced and nothing acts on it. Retrieval that fires on what the model generates is the active-retrieval
idea~\citep{jiang2023flare}, and naming before retrieving is the generate-then-retrieve
idea~\citep{gao2023hyde}; what is specific here is that the generated word fetches into the model's own key-value
cache, by the same index and the same slots as the request, and that the request is re-encoded over what arrives.

On the engine, \Llama{} at 128K reads about four fifths of what dense attention reads, and the verbatim handle
nearly as often as dense attention. \Mistral{} at 32K reads well below dense attention and below the harness, and
the gap to the harness is the far window. The engine's far window is always full (Section~\ref{sec:design:store});
at 32K its back-fill of the most recent far tokens covers about two thirds of the far region, so a same-format decoy
is resident from the hand-off in nearly every item, and when the fetched sentence arrives and the request is
re-encoded, the read goes to a resident decoy in most such rows, where the harness, whose far window held only
fetched units, read the fetched sentence in nearly all of them. Denying the far window the same-format decoys
recovers more than half of the gap, with the decoy reads gone and the remaining misses reading some other resident
token; turning the re-encode off changes little, so the re-encode neither causes nor cures the drop. The cost is the
far window's volume (Section~\ref{sec:design:store}) paid on the one read that competes on format rather than on
words; measuring this match with a far window that holds only fetched units remains future work, and the
\Mistral{} engine number is a property of the fill rather than of the model or the store. On \Llama{} at 128K the
back-fill covers a small part of the far region, few decoys are resident, and nearly every re-encoded row reads.

Native thinking does not replace the naming step. DeepSeek-R1-Distill-Llama-8B~\citep{deepseekr1}, asked the same
questions under the design with its own reasoning trace, rarely names the target and never reads the code, and its
verbatim read falls by a quarter once the trace is on; under dense attention the same model names the target about
as often as the base models do and reads the code less often. The effective procedure is the short one: name, then
answer. An extension of this procedure fetches the name's unit and glosses it into the request before encoding,
which in the harness reads about as well as naming does on both models, with and without a worked example, well
above its wrong-name controls.  We leave this for future work.

\begin{table}[htbp]
\centering\footnotesize
\caption{Matching on descriptions. Description-matched questions with three same-format decoys per document,
$n = 81$, exact seven-digit match. ``Harness'' is the PyTorch reference, whose far window holds only fetched units;
``engine'' is the \llamacpp{} implementation, whose far window is always full. ``Name first'' asks the model to name
the described thing before giving the code; ``re-encode'' is the roll-forward over the request and the generated
tail after the fetch.}
\label{tab:concept}
\begin{tabular}{@{}p{7.6cm}p{3.9cm}p{4.0cm}@{}}
\toprule
arm & \Mistral{} 32K & \Llama{} 128K \\
\midrule
\multicolumn{3}{@{}l}{\itshape naming, under dense attention} \\
the description asked directly (harness) & \rd{9}{81} & \rd{13}{81} \\
name first, then the code (harness) & \rd{56}{81} & \rd{61}{81} \\
name first (engine) & \rd{59}{81} & \rd{64}{81} \\
the thing named at all (harness) & \rd{58}{81} & \rd{64}{81} \\
given a correct name & & \rd{59}{59} \\
\addlinespace
\multicolumn{3}{@{}l}{\itshape the design, name first} \\
events off (harness) & \rd{0}{81} & \rd{0}{81} \\
event fetch, no re-encode (harness) & \rd{2}{81} & \rd{44}{81}, 44 of the 51 units fetched \\
event fetch and re-encode (harness; the far window holds fetched units only) & \rd{51}{81}, past a bar of 41 & not run \\
engine, the far window always full & \rd{36}{81} & \rd{50}{81}--\rd{51}{81} \\
\quad same, same-format decoys denied from the far window (oracle) & \rd{45}{81} & \\
\quad same, re-encode off & \rd{35}{81} & \\
the verbatim handle asked (engine; design / dense) & \rd{70}{81} / \rd{78}{81} & \rd{73}{81} / \rd{79}{81} \\
the thing named at all (engine; design / dense) & \rd{58}{81} / \rd{58}{81} & \rd{53}{81} / \rd{63}{81} \\
\addlinespace
\multicolumn{3}{@{}l}{\itshape why the engine's \Mistral{} number is the fill} \\
a same-format decoy resident from the hand-off & 73 of 81 items & few: the back-fill covers 16K of 123K far tokens \\
rows where the event fetched the unit and the request was re-encoded & the fetch loses to a resident decoy in 17 of 20 (harness: the fetched unit read in 50 of 51) & 42 of 43 read \\
\addlinespace
\multicolumn{3}{@{}l}{\itshape a native reasoning trace: DeepSeek-R1-Distill-Llama-8B, under the design / under dense attention} \\
names the target & \multicolumn{2}{l}{\rd{13}{81} / \rd{66}{81}} \\
reads the code & \multicolumn{2}{l}{\rd{0}{81} / \rd{43}{81}} \\
the verbatim read without / with the trace, under the design & \multicolumn{2}{l}{\rd{80}{81} / \rd{60}{81}} \\
\addlinespace
\multicolumn{3}{@{}l}{\itshape follow-up: the name's unit glossed into the request before encoding (harness; without / with a worked example)} \\
with the correct name & \rd{60}{81} / \rd{58}{81} & \rd{59}{81} / \rd{54}{81} \\
with a wrong name (control) & \rd{29}{81} / \rd{2}{81} & \rd{48}{81} / \rd{1}{81} \\
\bottomrule
\end{tabular}

\end{table}

\secbarrier

\section{Limitations}
\label{sec:limitations}

The first hop limits two-hop questions. They reach parity with dense attention on \Llama{} at 128K and stay well below it on \Mistral{} at 32K (Table~\ref{tab:twohop}). The
limit is the first hop, and in the engine it is the near window rather than the far window: at 32K the
bridge word is produced about as often with the needle resident as by the window-only floor, and far more often
under dense attention, so the near window alone loses most of the hop and the far window a little; at 128K the far
window's share grows. On \Llama{} the answer is right far more often than the bridge word is produced, so the
first-hop reading bounds \Mistral{}'s result rather than \Llama{}'s. Table~\ref{tab:misses} sorts the misses with the far seed off by cause: on \Mistral{} most are units resident without the bridge word being produced, and on
\Llama{} most were handles outside the seed, a class the far seed removed. A pointer-aware read, or a second
re-encode after the first hop's fetch, is a candidate remedy.  It also suggests that the design is best applied to models trained to a longer range; a 32K-trained attention window can be stretched only so far.

\begin{table}[htbp]
\centering\footnotesize
\caption{Two-hop misses by cause with the far seed off, $n = 100$ per row. The far seed removes the handle-outside-the-seed class on \Llama{}.}
\label{tab:misses}
\begin{tabular}{@{}llrp{9.4cm}@{}}
\toprule
model & context & misses & what the misses were \\
\midrule
\Mistral{} & 32K & 38 & 6 units not admitted; 23 resident without the bridge word; 9 resident with the word produced and misread \\
\Mistral{} & 128K & 45 & 11 not admitted; 25 resident without the bridge word; 4 wrong after a re-encode; 5 misread \\
\Llama{} & 32K & 26 & 14 handles outside the 32-word seed; 12 no bridge word \\
\Llama{} & 128K & 16 & 9 handles outside the seed; 7 no bridge word \\
\bottomrule
\end{tabular}

\end{table}

The always-full far window costs reads. The far window back-fills to 16,384 slots with the most recent
far tokens, and on reads that compete on format rather than on words that content costs: the description read
through naming on \Mistral{} loses nearly a third of its reads against a far window of fetched units alone
(Table~\ref{tab:concept}), and two of the three multi-key RULER tasks on \Mistral{} at 32K lose about twenty points to dense
attention (Table~\ref{tab:ruler}). Measuring the description match with a far window that holds only fetched units remains future work (Section~\ref{sec:concept}).

The RULER results are mixed. The design's aggregate is a point above dense attention on \Llama{} at 128K and eight points
below it on \Mistral{} at 32K (Table~\ref{tab:ruler}). The usual bar of 50 on every multi-key task is missed by three
points on the third multi-key task; the tasks that chain through content, variable tracking and the first
question-answering task, fall below dense attention on \Llama{}; the volume tasks, common and frequent words, fall
below it on \Mistral{}, where the window-unit flag recovers part of the loss (Table~\ref{tab:design}). The matched floor, a 24,576-token window with no retrieval, scores far below the design on
\Llama{} at 128K, so the aggregate comes from the far window rather than from the budget. The remaining controls the
RULER bar asks for, a random store and the published training-free systems InfLLM~\citep{xiao2024infllm} and
ReAttention~\citep{liu2025reattention}, were not run against the same items; and a needle that shares no word with
its question, the NoLiMa setting~\citep{modarressi2025nolima}, is outside what a lexical index can fetch without the
naming step of Section~\ref{sec:concept}.

Throughput below the trained range misses its target of parity with the stock engine. Prefill and steady decode each cost under ten percent there (Table~\ref{tab:perf}), from the copy that streams rolled-off keys and
values to the host and from the window-only prefill above 24.7K tokens.

The host store implemented in \llamacpp{} is unbounded and uncompressed, and holds about 130 GiB at one million tokens on \Mistral{}. This is a practical limit on serving an unbounded context. The compression of Section~\ref{sec:design:range} cuts the store four and a half times, and beyond that the remedy is a further tier: spill the store to disk and page units in as the index names them, which the word index already makes possible, since a unit is fetched by its words rather than scanned.

The scope is two 7--8B models of one architecture family, one item seed per row, and a 70B we measured only
under full attention in the harness. Every model is rotary and dense, and the design refuses multi-head latent
attention and linear-bias models by construction. Single-seed differences are read as parity, never as gains. The
residency rule is measured on short answers; tokens that regain attention late in a long reasoning
trace~\citep{zhang2025lazyeviction} are outside the measured regime, which is why the re-encode carries a per-turn
budget. The census of Section~\ref{sec:what} is over 2,048-token windows, and its sink share and far-content class
depend on that length. The order-at-write half of the plane account rests on the writer-reader weights and a
collapsed-position order question rather than on a write-time perturbation.

\secbarrier

\section{Discussion}
\label{sec:discussion}

We built the unbounded context for the straightforward use of a transformer model, one long document and a question at its end, but the concepts presented here extend to other uses. In this section we consider a few of the more fruitful directions.

The first direction is retrieval into the attention rather than into the prompt. Retrieval-augmented generation puts retrieved text into the prompt~\citep{lewis2020rag,ram2023incontextralm,shi2024replug,izacard2023atlas}: an external index
chooses passages, the passages are prepended as tokens, the model encodes them again at every request, and it reads
them as text. The design here augments the attention instead. A span is pre-encoded through the model once and its
keys and values are stored; the word index records its rare words; and when a request names one of them the span
enters the far window as keys and values at a slot, where the model reads it as it would have read it in place. The
measurement that licenses the first step is in Section~\ref{sec:values}: a forward pass over a unit's own tokens with
no preceding context reads every needle from 32K to one million tokens, at a bulk cost of a few hundredths of a nat
that a few hundred tokens of its own context removes (Table~\ref{tab:values}). So the context of the model need not be the
conversation. A library, a code base, or a user's past sessions can be encoded once, offline, and become far context
under the same index and the same far window, and the retrieved material then costs no prompt tokens, no re-encoding of the
passages, and no place at the front of the prompt where order effects apply. The serving systems that precompute
chunk caches for retrieval~\citep{gim2024promptcache,lu2025turborag,ma2025blockattention,yang2025ape,yang2025kvlink,yao2024cacheblend,jin2024ragcache,agarwal2025cachecraft,chan2025cag,li2026takenotes}
already build that machinery, most of them for prompt-scale contexts and with a fine-tune or a recompute halo that
restores attention across chunk boundaries, and knowledge bases have been written into attention as key and value
vectors through trained adapters~\citep{wang2025kblam} or a model trained for it~\citep{yang2024memory3}. What this
paper adds to that program is where and when a pre-encoded span reads on a stock model: at any slot inside the
trained range, degrading beyond about twice it; as the model's own keys and values rather than a synthesized
stand-in; and resident when the question's handle word is encoded. Whether a span encoded without its document reads
as well under competition as one captured from the window remains to be measured.

The second is that the retriever and the reader are decoupled. The word index is the retriever here, and it is a weak one on
purpose: whole rare words, exact match, scored by surprisal. It is enough for every planted row, because a planted
item has a handle, and the rows it loses on RULER are the two kinds a lexical index cannot serve: a question that
shares no rare word with its answer, and many spans that share the question's words. NoLiMa is the benchmark form of
the first~\citep{modarressi2025nolima}, needles with no literal overlap with the question, and the description match of Section~\ref{sec:concept} is our answer to it: the model names the thing, and the name is the key. A better
scorer is the answer to the second, and the far window is the place to measure one, since any ranking that produces units
can enter it by the same placement: a dense retriever's top units, a hybrid of words and vectors, or the model's own
keys as the key-similarity family uses them~\citep{xiao2024infllm,liu2024retrievalattention,chen2025retroinfer}. The
attention performs the final read in every case, so a far window holding only fetched units isolates the reader
from the retriever and is the natural next measurement.

The third is that timing is a rule for every retrieval system. The residency finding says more than fetch first. The
binding forms at the handle word's own query, and a unit that arrives later is unread until the question is encoded
over it. For a prompt-side pipeline that is the rule that retrieved text precedes the question, which practice already
follows. For a system that retrieves while it generates~\citep{jiang2023flare,su2024dragin,asai2024selfrag} it says
that the retrieved material must be encoded together with the question, which the prompt route pays for with the whole
prompt and the attention route pays for with one exact re-encode, a re-run of the request and the generated tail over what arrived, a few tens of milliseconds here (Table~\ref{tab:design}). An agent that calls a tool mid-generation and reads the result faces the same rule, and the event
mechanism is one way to meet it inside the cache. The cost side has a lesson as well: the gate on the generated word's
surprisal cut the re-encodes to a few per hundred generated tokens without losing a read, so the model's own uncertainty about a word
is an adequate trigger, as the active-retrieval line found on the prompt side.

The fourth is a memory that outlives the session. The host store is position-free and exact, so nothing in it depends
on the session that wrote it, and resuming a session from its store is a test the implementation has not run. A store
that persists is a memory of everything the model has read, with the index as its catalog, and the question it raises
is the question that usage-based eviction cannot answer: what to forget. On a document the answer was nothing, because no far
unit is used before its question. On a lifetime store the criterion is open; age, source and index demand are the
candidates, and the measurement to make is the same three-key discrimination under each. The costs are concrete. At
131 KB per token a terabyte of host memory holds about eight million tokens exact; at the 29 KB of the measured
28-plane four-bit rung it holds about 35 million; the gather moves a few thousand tokens per request at one million,
and the GPU working set is fixed at the near window, the far window and the pins. The memory wall moves from the device to the
interconnect, and the traffic across it is a few thousand cells per request, small against what the link carries.
Agent memories kept as keys and values rather than as text~\citep{pandey2026agentkvshift,packer2023memgpt} are the
same idea seen from the application side.

The fifth is the line between what the stock model does and what training adds. Every read here is lexical, a description does not
cross the window, and a second hop is work done while generating. Those are properties of the models as trained, and
the retrofit exposes them rather than changing them. Models trained to longer contexts~\citep{gao2025prolong,xu2025ultralong}
buy a longer window; what they buy for a far read is unmeasured here, and the census suggests the answer is the same
handful of reader heads at a longer reach. The more interesting question is the reverse one: a model fine-tuned with a far window present, so that it learns to read fetched units by what they mean rather than by the words they share with the
question, would move the description read into the retrieval tier. The trained-memory line~\citep{wu2022memorizing,mohtashami2023landmark,munkhdalai2024infiniattention}
has built such readers from scratch; whether one can be grafted onto a stock model with a few steps of training, while
the window and the sink stay as they are, is a question the engine makes testable.  More generally, Section~\ref{sec:geometry} presents a functional view of what attention does. A model and a training regime aimed at those functions, rather than left to discover them from a generic initialization, could serve them better; that is a direction for training rather than for serving.

The last direction is serving less of the cache. Half of the attention mass is routing within 64 tokens, a few mid-stack heads do the far
reading, and most of the rest is parked on the sink. The design ignores that: one store, one index, one far window, read by every layer alike. A design that kept the far window only for the layers that read it, or scheduled the gather by which
heads will fire, would cut the GPU cache below its present bound, and the head-level cache designs~\citep{xiao2024duoattention,tang2024razorattention,fu2025notallheads}
show the magnitude of the available saving. The selection itself stays lexical in such a design, since a content match does not make a far unit readable (Section~\ref{sec:descriptions}); the saving comes from serving the reads where they occur.

\secbarrier

\section{Related work}
\label{sec:related}

The design sits at the meeting point of several literatures: streaming attention and its sinks, caches that store
keys without a position, retrieval over the cache, eviction and salience rules, trained memories, the reuse of
precomputed caches for retrieval, the timing of a read, the structure of rotary attention, the heads that copy, and
retrieval that fires while a model generates. Each block below names the work our design stands on and states what we add to it.

\paragraph{Streaming windows and sinks.} StreamingLLM keeps a near window and the first tokens as attention sinks
and shows that removing the sinks is catastrophic~\citep{xiao2024streamingllm}; LM-Infinite reaches the same
window-plus-anchor shape through a $\Lambda$-shaped mask and a distance ceiling~\citep{han2024lminfinite}. The sinks
are the massive activations of \citet{sun2024massiveactivations}; \citet{gu2025attentionsink} trace when they emerge
in training, \citet{barbero2025firsttoken} read first-token attention as a control on over-mixing, and KVSink
preserves them under quantization~\citep{su2025kvsink}. SepLLM keeps separators as segment summaries~\citep{chen2025sepllm},
and Star Attention prefixes an anchor block to every block so that each has a sink to stand on~\citep{acharya2025starattention}.
What is specific here is where the sink must sit: at a slot inside the trained range, since at its true distance it
costs $+0.58$ and $+1.15$ nats by one million tokens, and the drift of a sink key moved in place about 8,100 times in
f16, which the engine fixes by re-placing the pinned prefix from the host. The census adds the sink's share of the
mass per layer and its reading as the softmax floor rather than a read.

\paragraph{Position-free storage and re-placement.} Storing keys without a position and rotating them at a new one
is the primitive of CachedAttention~\citep{gao2024cachedattention}, EPIC~\citep{hu2024epic},
MiniPIC~\citep{ordonez2026minipic}, LazyAttention~\citep{xia2026lazyattention} and Kamera~\citep{ma2026kamera}, and
CacheFocus re-positions cached document chunks into the trained length for retrieval~\citep{lee2025cachefocus}.
ReAttention selects far tokens without position and rotates the selected ones into a finite scope, training-free, on
these same models~\citep{liu2025reattention}; Extended Mind Transformers retrieve the model's own cached keys and
values by similarity and find that no position at all serves them best under rotary
embeddings~\citep{klett2024extendedmind}; Parallel Context Windows reuse one range of positions for several
windows~\citep{ratner2023pcw}. SelKV reports that re-embedding kept keys at contiguous positions can hurt a
long-context read~\citep{bouyahiaoui2026selkv}, which our tolerance measurement bounds. The length-extension
line moves the positions instead of the keys: interpolation~\citep{chen2023pi}, per-band scaling~\citep{peng2024yarn,ding2024longrope,shang2025longrope2},
grouped far positions~\citep{jin2024selfextend}, chunked remapping~\citep{an2024chunkllama}, a shift over the
under-trained tail~\citep{an2025string} and a clip of the slow components~\citep{li2026cope}; the trained extensions
of \citet{gao2025prolong} and \citet{xu2025ultralong} buy the range outright. The measurement added here is the
tolerance of the far match: any slot inside the trained range, a slow degradation to about twice it, failure at
distance zero and at the sink, and a GPU cache bounded by a word index rather than by the document.

\paragraph{Retrieval over the cache and sparse attention.} The training-free systems closest in shape keep a sink, a
window and far units in host memory and fetch units by a similarity between the query and a representative of the
unit: InfLLM~\citep{xiao2024infllm}, LongHeads~\citep{lu2024longheads}, RetrievalAttention and
RetroInfer~\citep{liu2024retrievalattention,chen2025retroinfer}, MagicPIG~\citep{chen2025magicpig},
PQCache~\citep{zhang2025pqcache}, ShadowKV~\citep{sun2024shadowkv}, ClusterKV~\citep{liu2024clusterkv},
TokenSelect~\citep{wu2025tokenselect}, InfiniGen~\citep{lee2024infinigen}, HeadInfer~\citep{luo2025headinfer} and
InfiniteHiP, which reaches three million tokens on one GPU~\citep{lee2025infinitehip}. Within the trained length,
Quest~\citep{tang2024quest}, MInference~\citep{jiang2024minference}, SparQ~\citep{ribar2024sparq},
Loki~\citep{singhania2024loki}, HashAttention~\citep{desai2025hashattention}, Squeezed
Attention~\citep{hooper2025squeezed}, TidalDecode~\citep{yang2024tidaldecode}, DuoAttention~\citep{xiao2024duoattention}
and RazorAttention~\citep{tang2024razorattention} select blocks or heads by key statistics at run time; Multipole
Attention and CEDAR score centroids or chunk summaries inside one softmax~\citep{hooper2025multipole,li2026cedar},
and vAttention bounds the error of such a selection~\citep{desai2026vattention}. EM-LLM segments the context into
events by surprise and retrieves them by similarity and contiguity~\citep{fountas2025emllm}, and S$^3$-Attention
keeps a CPU inverted index of learned sparse features and fuses BM25 matching for rare entities~\citep{ma2026s3attention},
the nearest lexical neighbor of the index here. Native Sparse Attention, MoBA and the sparse attention of
DeepSeek-V3.2 train the selection in~\citep{yuan2025nsa,lu2025moba,deepseek2025v32}. The retrofit here selects by
words rather than by keys, needs nothing trained, and places whole sentence units into a far window whose tolerance is measured.

\paragraph{Eviction, usage and salience.} Usage-scored stores run from the least-recently-used memory of
\citet{santoro2016mann} to heavy-hitter eviction~\citep{zhang2023h2o}, the persistence-of-importance hypothesis of
Scissorhands~\citep{liu2023scissorhands}, the observation window of SnapKV~\citep{li2024snapkv}, the layer budgets of
PyramidKV~\citep{cai2024pyramidkv}, TOVA~\citep{oren2024tova}, cascading caches~\citep{willette2024cascadekv} and
forgetting factors~\citep{jo2024a2sf}; learned retention gates follow~\citep{bui2025trimkv,bui2026makeeachtoken,yang2026indexmem},
and \citet{steele2026limits} and \citet{kumar2026page} measure where learned importance scoring stops working.
Attendre keeps evicted queries so that a key can still be attended by a query that arrives later~\citep{yang2024attendre},
the nearest precedent for the timing problem of Section~\ref{sec:residency}. Lexical and salience rules decide what
to retain in Transactional Attention~\citep{basu2026transactional}, on these same models, in SirLLM~\citep{yao2024sirllm},
SALT~\citep{mamo2026salt} and Pulsar Attention~\citep{sood2026pulsar}, TriRoute learns the routing~\citep{balashov2026triroute},
and BudgetMem filters chunks by lexical statistics on the prompt side~\citep{alla2025budgetmem}. The admission rule
here is a member of that family, and what we add to it is the demand bound at one million tokens and the finding that usage-at-exit eviction of a host store fails on a long document, because no far unit is used before its question.

\paragraph{Trained memories and compressed contexts.} Memorizing Transformers~\citep{wu2022memorizing}, Landmark
Attention~\citep{mohtashami2023landmark}, Unlimiformer~\citep{bertsch2023unlimiformer}, the Focused
Transformer~\citep{tworkowski2023fot}, LongMem~\citep{wang2023longmem}, MemLong~\citep{liu2024memlong},
Infini-attention~\citep{munkhdalai2024infiniattention}, Transformer-XL~\citep{dai2019transformerxl} and the $k$NN
language model~\citep{khandelwal2020knnlm} train for a memory; a reproduction of Infini-attention reports that its
compressed memory struggles to retrieve from earlier segments~\citep{nguyen2024infinifailed}. Compression of the
context into trained tokens or states~\citep{rae2020compressive,mu2023gist,chevalier2023autocompressor,ge2024icae,zhang2024beacon}
and latent memory pools~\citep{wang2024memoryllm} trade the exact history for a smaller one, and RetroLM retrieves
cache pages with a trained page retriever~\citep{luo2025retrolm}. The retrofit here trains nothing and keeps the far
history exact; its measurement that a synthesized key or value reads nothing while a self-consistent forward over the
unit's own text reads \rd{14}{14} at one million tokens is the reason the store holds the model's own cells.

\paragraph{Reusing precomputed caches for retrieval.} Prompt Cache reuses the keys and values of prompt modules at
schema-assigned positions~\citep{gim2024promptcache}, and the retrieval-augmented serving line precomputes a cache per
chunk and concatenates the chunks a request retrieves: TurboRAG~\citep{lu2025turborag},
Block-Attention~\citep{ma2025blockattention} and KVLink~\citep{yang2025kvlink} fine-tune the model to the
independently encoded chunks, APE aligns them without training through a shared prefix, a scale and a temperature~\citep{yang2025ape}, and CacheBlend~\citep{yao2024cacheblend}, Cache-Craft~\citep{agarwal2025cachecraft},
EPIC~\citep{hu2024epic}, Star Attention~\citep{acharya2025starattention} and \citet{mao2026contiguity} recompute a
small set of tokens so that reused chunks read correctly, while \citet{tachibana2026recompute} combine a
concatenation-aware fine-tune with recomputation and report the one RULER number in that line past 100K tokens.
Cache-augmented generation preloads a whole knowledge base as a cache inside the window~\citep{chan2025cag}, REFRAG
feeds compressed chunk embeddings to the decoder and expands the ones a policy selects~\citep{lin2025refrag},
\citet{li2026takenotes} precompile chunks and re-rotate their keys before splicing them in, and agent memories are
kept as caches rather than as text~\citep{pandey2026agentkvshift}. Serving systems supply the tiers such caches live
in~\citep{liu2024cachegen,qin2025mooncake,liu2025lmcache,zheng2024sglang,gao2025hcache}, and KVShareArena benchmarks
the reuse~\citep{shi2026kvsharearena}. What is added here is the unit-level result at range on a stock model: a span
rebuilt from its own tokens with no preceding context reads \rd{14}{14} at one million tokens where a higher-fidelity
synthesized key reads nothing, so a self-consistent forward pass outperforms a better map, and the slot it is placed at may be
anywhere inside the trained range.

\paragraph{Residency and query visibility.} Lost in the Middle moves the question to show that content encoded with
the question visible is read better~\citep{liu2024lostmiddle}; \citet{luo2026queryvis} show that compression
rankings invert with the scorer's query visibility, WnW that prefill and decode rankings overlap
weakly~\citep{yao2026wnw}, and KVzip, Lookahead Q-Cache, DapQ and Expected Attention estimate the queries a cache will
face before they arrive~\citep{kim2025kvzip,wang2025laq,tian2026dapq,devoto2025expectedattention}. LazyEviction shows
tokens regaining attention late in long reasoning traces~\citep{zhang2025lazyeviction}, Retrospective Sparse
Attention revises earlier outputs when later keys arrive~\citep{choi2025retroattention}, and \citet{okamoto2026encodedearly}
find a signal encoded early and acted on late. The residency measurements here hold content, order and attention mass
fixed and vary only the moment a far unit becomes visible, which locates the binding at the handle word's own query
under competition and motivates the exact re-encode at an event, with a per-turn budget for the regime LazyEviction
describes.

\paragraph{Rotary planes and key compression.} The semantic and positional roles of rotary
frequencies~\citep{su2021roformer,barbero2024roundandround,hong2024tokendistance,peng2024yarn} are the ground of this
paper's plane account; \citet{chiang2025dimension} find that long-distance retrieval rides on the lowest-frequency
dimensions, \citet{oka2026frequencybands} locate a high-norm band of frequencies whose place depends on the base,
\citet{men2024baserope} bound the context length by the base, and \citet{du2026ropeneither} prove what rotary
attention can and cannot distinguish at long range. The analytic per-wavelength window of
ATFlash~\citep{hayashi2026atflash} is the model-card plane cut used here; the frequency-selecting compressions
EliteKV, RAP, FASA and KV-Latent~\citep{zhou2025elitekv,xin2026rap,wang2026fasa,shi2025kvlatent} and the key-channel
pruning of ThinK, LeanK, Prism and Thin Keys, Full Values~\citep{xu2025think,zhang2025leank,wang2026prism,yao2026thinkeys}
cut the same dimensions by a learned or measured score; KIVI, KVQuant, NSNQuant and TriAttention quantize or
compress the cache~\citep{liu2024kivi,hooper2024kvquant,son2025nsnquant,mao2026triattention}. The plane results here
are the measured cliff on multi-candidate reads between 22 and 15 planes, the fact that loss cannot see it, and the
sink's destruction when only fast planes are kept; the design refuses multi-head latent
attention~\citep{deepseekv2} and linear-bias schemes~\citep{press2022alibi}, which have no per-plane rotation to place.

\paragraph{Heads, induction and copying.} Head-role studies~\citep{voita2019heads,clark2019bert,zheng2024headsurvey}
and induction heads~\citep{elhage2021framework,olsson2022induction,bietti2023birth} are the classic accounts; PAPA
asks how much attention is needed at all~\citep{hassid2022papa}, and \citet{chang2025bigram} map the bigram
subnetwork that next-token prediction rests on. Retrieval heads~\citep{wu2024retrievalhead,zhang2025qrhead,bayram2026retrievalheads,lee2025seal,fu2025notallheads},
binding identifiers~\citep{feng2024binding}, information flow routes~\citep{ferrando2024routes} and factual
recall~\citep{geva2023dissecting} locate the copy, Future Lens reads its carry~\citep{pal2023futurelens}, copy
suppression is its counterweight~\citep{mcdougall2023copysuppression}, and \citet{hayes2026programs} and
\citet{huang2026decompiling} synthesize a program per head or per model. The dual-route model separates a verbatim
from a semantic induction route by ablation~\citep{feucht2025dualroute}; \citet{dangelo2026ngram} show on synthetic
data that induction weights a context match exponentially by its overlap, and the generalized induction head of
\citet{kim2025gim} adds fuzzy matching to the count. The aligned-match result of Section~\ref{sec:what} is the
deployment-scale form of that finding, on a 7B model reading prose. The census here is a pooled mass accounting over
every head with the sinks reported separately; its text-only chooser is of the suffix kind that infini-gram scales to
a trillion tokens~\citep{liu2024infinigram}; and it stays behavioral: no head is ablated and no head is replaced.

\paragraph{Values and words.} That a multi-token word is assembled early in the residual stream is the finding of
\citet{feucht2024erasure}, \citet{kaplan2025tokenswords}, \citet{kamoda2025weightdetok} and
\citet{busigin2026wordfactory}, and the staged picture of inference of \citet{lad2024stages}; the lenses that read
the residual~\citep{belrose2023tunedlens,ghandeharioun2024patchscopes} and the additive account of
recall~\citep{chughtai2024summing} are the instruments. On the cache side, KV Prediction approximates keys and values
for a faster first token~\citep{horton2024kvprediction} and AsymKV keeps the values lossless because they tolerate
less~\citep{cui2025asymkv}. What we add is the serving consequence: a synthesized value destroys the read
while a synthesized key only misaddresses it, and a word's assembled vector is order-coded and readable through the
embedding table where the logit lens is not.

\paragraph{Retrieval during generation.} FLARE retrieves when a generated span is uncertain~\citep{jiang2023flare},
DRAGIN and SeaKR read the model's own information need~\citep{su2024dragin,yao2024seakr}, Self-RAG trains reflection
tokens that decide to retrieve~\citep{asai2024selfrag}, Adaptive-RAG routes by question complexity~\citep{jeong2024adaptiverag},
and \citet{zhang2026prefetch} prefetch what a generation will ask for. IRCoT interleaves retrieval with reasoning
steps~\citep{trivedi2023ircot}, Self-Ask decomposes the question~\citep{press2023selfask}, and HyDE, Query2doc and
query rewriting generate before they retrieve~\citep{gao2023hyde,wang2023query2doc,ma2023rewrite}; MemoRAG drafts
clues from a global memory~\citep{qian2025memorag}, and MemGPT and ReadAgent page text in and out of the
context~\citep{packer2023memgpt,lee2024readagent}. Retrieval augmentation itself~\citep{lewis2020rag,borgeaud2022retro,shi2024replug,ram2023incontextralm}
and its comparison with long-context models~\citep{li2024selfroute,jiang2024longrag} frame the question Section~\ref{sec:discussion}
takes up. The events and the description match are that idea inside the model's own cache: a generated rare word fetches
its unit by the same index and into the same slots as the request, and the re-encode is exact.

\paragraph{Evaluation.} The needle protocol is Kamradt's~\citep{kamradt2023niah}; RULER supplies the generator, the
tasks and the metric~\citep{hsieh2024ruler}; LongBench, $\infty$Bench, HELMET, LV-Eval and BABILong extend the
evaluation to realistic tasks and to millions of tokens~\citep{bai2024longbench,bai2025longbenchv2,zhang2024infinitebench,yen2025helmet,yuan2024lveval,kuratov2024babilong},
Needle Threading follows chains through near-million-token haystacks~\citep{roberts2025needlethreading}, and NoLiMa
removes the literal match between question and needle~\citep{modarressi2025nolima}, the test a lexical index must
eventually face. Long-range loss is known to be a poor guide to retrieval: \citet{sun2021longrange} found long-range
models barely use far context except for copy-like tokens, \citet{fang2025longppl} show perplexity missing
long-context ability, and retrieval helps the rare tokens that bulk loss barely weighs~\citep{nishida2025longtail},
which is why every store change here is validated on three-key discrimination rather than on loss. Prompt
compression~\citep{jiang2024longllmlingua,pan2024llmlingua2,xu2024recomp,cheng2024xrag} is the prompt-side
alternative to keeping the far text exact.

\secbarrier

\section{Conclusion}
\label{sec:conclusion}

This paper began by examining the geometry and function of attention in a stock rotary model. A far key is
addressed through its slowest rotary planes, which is why it can sit at any position inside the trained range and
still be found. The order of the tokens inside a span is fixed when their keys are written, so a span can be moved
after the fact without losing it. A far read is a copy: a few heads in the middle of the stack find an earlier
occurrence of the current word and copy what followed it, which is why a far span is reached by its words rather
than by its meaning. When several spans compete, the one that is read is the one present at the moment the
question's handle word is encoded, so a fetched span must arrive before the question and stay until the answer is
produced. The sink and the anchors hold the softmax floor only while they sit inside the trained range.

A stock decoder with those facts applied reads its whole history through
a bounded GPU cache, on the checkpoint and the runtime already in use, at one million tokens on one device, with
every setting but one read off the model card.  The far window that makes needle reads perfect is always full, and its
back-fill is what the one read that competes on format rather than on words pays for; a far window of fetched units alone
remains to be measured. The first hop of a two-hop question on a 7B model is lost by the near window before any
retrieval begins, which points at a second re-encode after the first hop's fetch, or at a read that follows a pointer.
A description is no address until the model has turned it into a word, and the model can do that in its own output:
matching on descriptions through the model's own naming is the start of a retrieval loop in which the model writes its own queries. Beyond the host horizon
a unit's keys can be rebuilt from its text, so the history need not even be stored as keys, and anything the model
has once encoded can enter its attention the same way, which is the retrieval Section~\ref{sec:discussion} describes. The deepest open question
is the one the census raised: half of attention is routing, a few heads do the reading, and the rest is mass parked
on anchors. A serving design that knows which heads read, and what they read by, could reduce the GPU cache further.

\secbarrier

\bibliographystyle{plainnat}
{\small\bibliography{refs}}

\end{document}